\documentclass[preprints,article,submit,pdftex,moreauthors,onecolumn]{Definitions/mdpi}

\usepackage[utf8]{inputenc}
\usepackage[T1]{fontenc}

\usepackage{amsmath,amssymb,amsfonts}
\usepackage{mathtools}

\usepackage{graphicx}
\usepackage{booktabs}

\usepackage{algorithm}
\usepackage{algpseudocode}
\algrenewcommand\algorithmicrequire{\textbf{Input:}}
\algrenewcommand\algorithmicensure{\textbf{Output:}}

\firstpage{1}
\pubvolume{1}
\issuenum{1}
\articlenumber{0}
\pubyear{2026}
\copyrightyear{2025}
\datereceived{}
\dateaccepted{}
\datepublished{}

\Title{Camera Pose Revisited}
\Author{Władysław Skarbek $^{1}$, Michał Salomonowicz $^{2}$ and Michał   Król $^{3}$}

\address{%
$^{1}$ Warsaw University of Technology, Poland; wladyslaw.skarbek@pw.edu.pl\\
$^{2}$ Warsaw University of Technology, Poland; michal.salamonowicz.dokt@pw.edu.pl\\
$^{3}$ Independent Researcher; mkrol2k@gmail.com}

\corres{Correspondence: wladyslaw.skarbek@pw.edu.pl}

\newcommand{\SO}[1]{\mathrm{SO}(#1)}
\newcommand{\tr}{\operatorname{tr}}

\newcommand{\kernn}[1]{\kern#1pt}

\newcommand{\bb}[1]{\mathbb{#1}}
\newcommand{\mbm}[1]{\mathbf{#1}}
\newcommand{\cl}[1]{\mathcal{#1}}
\newcommand{\tp}[1]{{#1}^{\intercal}}

\newcommand{\inv}[1]{\in\bb{R}^{#1}}
\newcommand{\inm}[2]{\in\bb{R}^{#1\times#2}}

\newcommand{\ds}{\displaystyle}

\newcommand{\eqd}{\doteq}

\newcommand{\lra}{\longrightarrow}

\newcommand{\Lra}{\Longrightarrow}

\newcommand{\xeq}[1]{\overset{#1}{=}}
\newcommand{\ov}[1]{\overline{#1}}
\newcommand{\diag}[1]{\mathtt{diag}\left[#1\right]}

\newif\ifarxiv
\arxivtrue   

\begin{document}
\maketitle
\nolinenumbers

\ifarxiv
\begin{center}
\textbf{Abstract}
\end{center}
\begin{abstract}\relax

Estimating the position and orientation of a camera with respect to an observed scene is one of the central problems in computer vision, particularly in the context of camera calibration and multi-sensor systems. This paper addresses the planar Perspective--$n$--Point problem, with special emphasis on the initial estimation of the pose of a calibration object. As a solution, we propose the \texttt{PnP-ProCay78} algorithm, which combines the classical quadratic formulation of the reconstruction error with a Cayley parameterization of rotations and least-squares optimization. The key component of the method is a deterministic selection of starting points based on an analysis of the reconstruction error for two canonical vectors, allowing costly solution-space search procedures to be avoided. Experimental validation is performed using data acquired also from high-resolution RGB cameras and very low-resolution thermal cameras in an integrated RGB--IR setup. The results demonstrate that the proposed algorithm achieves practically the same projection accuracy as optimal \texttt{SQPnP} and slightly higher than \texttt{IPPE}, both prominent \texttt{PnP-OpenCV} procedures. However, \texttt{PnP-ProCay78} maintains a significantly simpler algorithmic structure. Moreover, the analysis of optimization trajectories in Cayley space provides an intuitive insight into the convergence process, making the method attractive also from a didactic perspective. Unlike existing PnP solvers, the proposed \texttt{PnP-ProCay78} algorithm combines projection error minimization with an analytically eliminated reconstruction-error surrogate for translation, yielding a hybrid cost formulation that is both geometrically transparent and computationally efficient.
\end{abstract}
\else
\abstract{
Estimating the position and orientation of a camera with respect to an observed scene is one of the central problems in computer vision, particularly in the context of camera calibration and multi-sensor systems. This paper addresses the planar Perspective--$n$--Point problem, with special emphasis on the initial estimation of the pose of a calibration object. As a solution, we propose the \texttt{PnP-ProCay78} algorithm, which combines the classical quadratic formulation of the reconstruction error with a Cayley parameterization of rotations and least-squares optimization. The key component of the method is a deterministic selection of starting points based on an analysis of the reconstruction error for two canonical vectors, allowing costly solution-space search procedures to be avoided. Experimental validation is performed using data acquired also from high-resolution RGB cameras and very low-resolution thermal cameras in an integrated RGB--IR setup. The results demonstrate that the proposed algorithm achieves practically the same projection accuracy as optimal \texttt{SQPnP} and slightly higher than \texttt{IPPE}, both prominent \texttt{PnP-OpenCV} procedures. However, \texttt{PnP-ProCay78} maintains a significantly simpler algorithmic structure. Moreover, the analysis of optimization trajectories in Cayley space provides an intuitive insight into the convergence process, making the method attractive also from a didactic perspective. Unlike existing PnP solvers, the proposed \texttt{PnP-ProCay78} algorithm combines projection error minimization with an analytically eliminated reconstruction-error surrogate for translation, yielding a hybrid cost formulation that is both geometrically transparent and computationally efficient.
}
\fi

\ifarxiv
\else
\fi

\section{Introduction}
 
Estimating the position and orientation of a camera with respect to an observed scene is one of the fundamental problems in computer vision and photogrammetry \cite{Hartley2004,Ma2004}. This task, commonly referred to as \emph{camera pose estimation}, constitutes a key component of many image processing systems, including three-dimensional reconstruction, augmented reality, mobile robotics, visual navigation systems, and camera calibration procedures \cite{Szeliski2010}. In the classical formulation, the problem reduces to determining a rigid (isometric) transformation  between the camera coordinate frame and the scene reference frame based on known correspondences between three-dimensional points and their two-dimensional image projections.

One of the most frequently studied variants of this problem is the \emph{Perspective--n--Point} (PnP) problem, in which the intrinsic parameters of the camera are assumed to be known, and the goal is to estimate its position and orientation from a finite set of 3D--2D correspondences. Despite decades of research and the availability of numerous algorithms with well-established theoretical properties, the PnP problem remains an active area of investigation. This is largely due to the inherent trade-offs between numerical stability, estimation accuracy, and computational complexity \cite{Lepetit2014}.


In practical library implementations, algorithms based on specialized optimization procedures or elaborate algebraic constructions currently dominate. While such methods often achieve excellent reconstruction accuracy, their internal structure can be difficult to analyze geometrically and to interpret intuitively. As a consequence, their transparency and pedagogical value are frequently limited. We observed this also in the case of the excellent state-of-the-art algorithm
 \texttt{OpenCV-SQPnP} of Terzakis and Lourakis \cite{Terzakis2020} which inspired us to make this research.

In this work, we adopt the perspective of revisiting the camera pose estimation problem, focusing not only on the final value of the cost function but also on the structure of the solution space, the choice of rotation parameterization, and the behavior of the optimization process itself. In particular, we consider representing camera orientation in Cayley space, a classical concept originating from early studies on the theory of rotations \cite{Cayley1846,Barfoot2017}, which exhibits favorable properties in the context of nonlinear optimization in contrast to other approaches to rotation parametrization  using quaternions or exponential maps \cite{Grassia1998}. For a more detailed discussion of the Cayley representation, including historical and terminological issues, see Appendix~\ref{app:cayley}.

Based on this formulation, we propose the \texttt{PnP-ProCay78} algorithm, which casts the camera pose estimation task as a least-squares problem solved using standard nonlinear optimization techniques. The proposed approach enables competitive camera pose estimation while maintaining a clear algorithmic structure and an intuitive geometric interpretation, making it attractive both for practical applications and for academic teaching.

\section{Main Contributions}

This paper presents a re-examination of the problem of camera pose and orientation estimation, with particular emphasis on the structure of the solution space, the choice of rotation representation, and the transparency of the optimization process. The main contributions of this work can be summarized as follows:

\begin{itemize}
  \item A coherent and geometrically interpretable formulation of the camera pose estimation problem is proposed based on the Cayley parametrization. This representation eliminates the need for explicitly enforcing positive determinant and orthogonality constraints on the rotation matrix and enables a direct analysis of optimization trajectories in the parameter space.

  \item The \texttt{PnP-ProCay78} algorithm is developed, in which the \emph{Perspective--$n$--Point} problem is formulated and solved as a nonlinear least-squares task using standard optimization methods, without resorting to specialized algebraic solvers or procedures dedicated exclusively to this problem.

  \item An initialization strategy based on two antipodal starting trajectories in the Cayley space is introduced. This significantly increases the robustness of the algorithm to unfavorable initial conditions and enables reliable convergence to the global minimum of the cost function over a wide range of geometric configurations.

  \item An alternative interpretation of the Cayley space is presented by decomposing the state vector into a directional component and a norm component. This leads to an intuitive visualization of optimization trajectories in terms of projections into the unit disk and distances from the origin. Such a representation proves particularly useful for geometric analysis and academic teaching.

  \item An experimental evaluation of the proposed approach is conducted in comparison with reference algorithms included in \texttt{OpenCV} library, demonstrating practically the same or higher estimation accuracy while maintaining a simpler algorithmic structure and full compatibility with commonly used geometric models and optimization tools.

  \item In the context of multisensor vision systems, a preliminary synchronous analysis of the optimization convergence is presented for a pair of RGB and thermal cameras operating within a single hardware setup. Despite substantial differences in image resolution and point localization accuracy between the two sensors, the \texttt{PnP-ProCay78} algorithm enables a consistent analysis of optimization trajectories within the same parameter space. At this stage, the analysis is visual and exploratory in nature, serving as a starting point for further research on convergence assessment in RGB--IR systems.
\end{itemize}

The presented results indicate that an appropriate choice of rotation representation combined with a conscious use of standard nonlinear optimization techniques makes it possible to construct a camera pose estimation algorithm that achieves high numerical quality while retaining conceptual clarity and strong didactic value, and that naturally fits within the framework of modern integrated vision systems.

\section{Related Work}

The development of algorithms for camera position and orientation estimation has progressed in parallel with advances in geometric modeling of the relationships between three-dimensional scenes and their image projections \cite{Lepetit2014}.

The earliest approaches to the Perspective--n--Point problem
and the camera pose estimation problems were comprehensively described in the classical monographs by Hartley and Zisserman \cite{Hartley2004} as well as by Ma et al.~\cite{Ma2004}, which to this day constitute the theoretical foundation for most contemporary algorithms.

The earliest approaches to the \emph{Perspective--$n$--Point} problem were primarily based on algebraic solutions, often relying on minimal point configurations. The classification of solutions to the P3P problem presented by Gao et al.~\cite{Gao2003, Kneip2011} together with the introduction of robust estimation techniques such as RANSAC proposed by Fischler and Bolles \cite{FischlerBolles1981}, enabled practical applications of camera pose estimation. However, these methods exhibited sensitivity to measurement noise and to geometric configurations close to degeneracy.

With the increasing availability of point correspondences, research shifted toward methods exploiting redundant data and formulating the PnP problem as an optimization task. A significant milestone in this direction was the EPnP algorithm proposed by Lepetit, Moreno-Noguer, and Fua \cite{Lepetit2009}, which allowed efficient processing of large point sets while maintaining low computational complexity. Subsequent works, such as the Direct Least Squares (DLS) method introduced by Hesch and Roumeliotis \cite{Hesch2011}, focused on improving numerical stability and estimation accuracy.

In parallel, approaches tailored to scenes with specific geometric structures—most notably planar as in work of Sturm and Maybank \cite{SturmMaybank1999}, frequently encountered in calibration tasks—were developed. In this context, the seminal works of Zhang \cite{Zhang2000} played a central role, followed by methods exploiting homography-based formulations and the separation of rotation and translation estimation. Among these, the IPPE algorithms proposed by Collins and Bartoli \cite{Collins2014} have found widespread use in practical library implementations.

In recent years, particular attention has been paid to algorithms that combine high estimation accuracy with robustness to near-degenerate configurations. A representative example is the SQPnP algorithm introduced by Terzakis and Lourakis \cite{Terzakis2020}, which employs sequential quadratic programming to compute a globally optimal solution. While such methods achieve excellent numerical performance, they rely on complex, highly specialized optimization procedures, which complicates geometric analysis and interpretation of the estimation process.

Regardless of the specific PnP algorithm employed, the representation of three-dimensional rotations remains a critical issue. Commonly used representations include Euler angles, quaternions, and Lie algebra–based formulations, whose properties have been extensively surveyed, among others, by Shuster \cite{Shuster1993}. The Cayley representation, originally introduced in the classical work of Arthur Cayley \cite{Cayley1846}, has so far been used relatively rarely in practical camera pose estimation algorithms, despite its favorable properties in the context of nonlinear optimization.

In the context of modern multi-sensor vision systems, increasing attention has also been devoted to camera pose estimation problems involving sensors with significantly different characteristics, such as RGB and thermal cameras. Studies on calibration and synchronization of such systems, including works by Prakash et al.~\cite{Prakash2019} and Lag\"uela et al.~\cite{Laguela2016}, indicate that differences in resolution and data quality have a substantial impact on the behavior of estimation algorithms. Complementary analyses of thermal mapping in multi-sensor systems have been presented, among others, by Borrmann et al.~\cite{Borrmann2014}.

Against this background, the approach proposed in the present work fits within the class of nonlinear optimization–based methods, while distinguishing itself through a deliberate choice of the Cayley parameterization and a strong emphasis on geometric analysis and interpretability of the estimation process. These aspects are particularly relevant in the context of multi-sensor systems and educational applications.

An important perspective that emerges when revisiting classical PnP algorithms is the distinction between the application-level performance metrics and the internal error (loss) function optimized by a given method. In practical vision systems, the quality of pose estimation is ultimately assessed using the projection error measured in the projection plane or reprojection error measured in the sensor plane.
In this work, projection error metrics are used for optimization, whereas reprojection error is employed solely for presenting results in pixel units.

However, most efficient PnP solvers do not minimize the projection error directly, but instead rely on surrogate cost functions that admit more favorable analytical or computational properties like reconstruction error. They are defined, for example, in \cite{Terzakis2020} as follows: for a set of  point correspondences $(P_i, p_i)$, i.e. for world object point $P_i$ and its projection point $p_i$, respectively, while error of pose $(R,t)$ is evaluated:
\begin{itemize}
\item \emph{Projection error}:
$$
\cl{E}_0^2 \eqd \sum_i\left\|p_i-\frac{(RP_i+t)}{(RP_i+t)_z}\right\|^2\,.
$$
\item \emph{Reconstruction error}:
$$
\cl{E}^2 \eqd \sum_i\left\|p_i\cdot(RP_i+t)_z-(RP_i+t)\right\|^2\,.
$$
\end{itemize}

This separation between the target metric and its optimization surrogate closely resembles a paradigm well known from modern machine learning, where task-level metrics are often replaced by differentiable or convex proxy losses. In the context of PnP, such surrogates include algebraic errors, quadratic reconstruction errors, or linearized residuals, which enable closed-form solutions
or fast convergence of iterative solvers.

\begin{table}[H]
\centering
\caption{Relationship between the application-level metric (projection error)
and the optimization surrogate used in selected PnP algorithms.}
\label{tab:pnp-costs}
\small
\begin{tabular}{p{2.8cm} p{3.4cm} p{4.4cm} p{3.6cm}}
\toprule
\textbf{Method} &
\textbf{Primary metric} &
\textbf{Optimization surrogate} &
\textbf{Remarks} \\
\midrule

EPnP \cite{Lepetit2009} &
Projection error &
Linearized control point fit using Gauss-Newton procedure &
Efficient for large $n$, moderate numerical stability \\

DLS \cite{Hesch2011} &
Projection error &
Polynomial least-squares formulation &
Improved numerical robustness; higher algebraic complexity \\

IPPE \cite{Collins2014} &
Projection error (planar scenes) &
Homography-based analytical solution  with hidden loss surrogates in standard procedures&
Specialized for planar targets; dual-solution ambiguity \\

SQPnP \cite{Terzakis2020} &
Projection error &
Quadratic reconstruction error
$\;\mathbf{r}^\top \Omega \mathbf{r}$ &
Globally optimal rotation via constrained quadratic programming \\

\textbf{PnP-ProCay78} &
Projection error (planar scenes) &
Closed formula for translation $t$ elimination +
projection residuals for rotation optimization &
Hybrid surrogate–metric design; Cayley parameterization enables
low-dimensional optimization and direct visualization\\

\bottomrule
\end{tabular}
\end{table}

For instance, while methods such as EPnP formulate pose estimation using algebraic constraints derived from control points, they do not explicitly optimize the projection error. 
However, to fit the control points, a Gauss--Newton optimizer is used.
Plane-based approaches such as IPPE exploit homography properties giving the closed formula solutions with respect to reference functions like \texttt{svd, rank}. 
This means that surrogate loss functions are implicitly embedded in matrix diagonalization procedures like finding singular values in \texttt{svd}. More recent methods, including DLS and SQPnP, explicitly introduce cost functions that approximate the projection error while remaining amenable to efficient global or near-global optimization. In table \ref{tab:pnp-costs} we show the comparison of selected methods with respect to this research aspect.

In this work, we follow this general principle by adopting the quadratic reconstruction error to get only closed formula for translation $t$ elimination while minimizing the projection error.

The proposed PnP-ProCay78 algorithm can thus be interpreted as a geometrically informed optimization of a well-chosen surrogate loss, whose minima are strongly correlated with low projection error in practice.

\section{Projective Models in Camera Calibration}

In this section, we introduce projective models and geometric concepts that form the foundation for the subsequent formulation of the camera pose identification problem. The presented definitions and properties are general in nature and are not yet tied to a specific rotation parameterization or a particular optimization algorithm. Their purpose is to establish a coherent conceptual framework that will be employed in the following sections devoted to the \texttt{PnP-ProCay78} algorithm.

\subsection{Matrix Formulation of the Pinhole Camera Model}

We begin with the standard definition of perspective projection used in the pinhole camera model. For clarity, we explicitly introduce both the input data and the unknown quantities, adopting a matrix notation that allows for a compact representation of subsequent definitions and properties.

\begin{itemize}
\item {\em Given:}
\begin{itemize}
\item $P\inm{n}{3}$ — coordinate vectors in the object reference frame defined for $n$ object points (for example points of a calibration target), expressed in physical units of the scene.
\item $p\inm{n}{2}$ — coordinate vectors in the optical reference frame corresponding to normalized projections of these points onto the plane $z=1$ (the $z$-coordinate is omitted).
\end{itemize}
\item {\em Unknown:}
\begin{itemize}
\item $R\inm{3}{3}$ — an orthonormal rotation matrix mapping the object frame to the optical frame,
\item $t\inv{3}$ — a translation vector specifying the position of the object frame origin expressed in the optical frame, with coordinates given in physical units of the scene.
\end{itemize}
\end{itemize}

A key element of the subsequent analysis is the definition of the reconstruction error, which measures the consistency between the projected scene points and their observations on the image plane. 
We adopt a row-wise and vectorized formulation that enables treating the camera pose estimation problem as a least-squares task. This notation is consistent with that used by the authors of the \texttt{SQPnP} algorithm~\cite{Terzakis2020}.

{\em Vectorized reconstruction error in row-wise form:}
\begin{itemize}\label{def-recon-error}
\item Object points expressed in the camera coordinate system:
\[
Q \eqd PR^\top + \mbm{1}_n t^\top .
\]
\item Depth vector of the object points: $Q_z\inv{n}$, where $Q=[Q_x, Q_y, Q_z]$.
\item Reconstruction error matrix and vector:
\[
E \doteq p \odot (Q_z \mbm{1}_2^\top) - [Q_x, Q_y]
\lra \vec{E} \eqd E.\mathrm{reshape}(2n).
\]
\end{itemize}

The above definition of the reconstruction error naturally leads to the question of its algebraic structure. In particular, we are interested in the possibility of eliminating the translation vector and reducing the problem to a form that depends solely on the rotation parameters. In this context, the following theorem plays a crucial role and forms the basis of many modern algorithms for solving the PnP problem.

\begin{Theorem}[On the stationary formula for the quadratic reconstruction error~\cite{Terzakis2020}]
\label{th:qform-greeks}
At stationary points with respect to the translation vector $t$, the reconstruction error of perspective projections $p_i\inv{3}$, $(p_i)_z=1$, equivalently the reconstruction error on the 3D scene side with respect to points $P_i\inv{3}$, given by
\[
\mathcal{E}^2 \eqd \frac{1}{2}\sum_i \|p_i (RP_i + t)_z - (RP_i + t)\|^2,
\]
can be expressed as a quadratic form for a certain matrix $\Omega\inm{9}{9}$ with respect to the variable $r\doteq \vec{R}$, $r\inv{9}$, i.e.,
\[
\mathcal{E}^2 = \frac{1}{2} r^\top \Omega r,\qquad t = \cl{T} r,
\]
where
\[
\begin{array}{l}\ds
\Omega \doteq \sum_{i=1}^n (\cl{P}_i+\cl{T})^{\top}\cl{Q}_i(\cl{P}_i+\cl{T}), \quad
\cl{Q}_i \doteq \left(p_i\mbm{e}_3^{\top}-\mbm{I}_3\right)^{\top}
\left(p_i\mbm{e}_3^{\top}-\mbm{I}_3\right),\\[6pt]\ds
\cl{T} \doteq -\left(\sum_{i=1}^n\cl{Q}_i\right)^{-1}
\left(\sum_{i=1}^n\cl{Q}_i\cl{P}_i\right),\\[15pt]\ds
\cl{P}_i \doteq
\begin{bmatrix}
P_i^{\top} & \mbm{0}_3^{\top} & \mbm{0}_3^{\top}\\
\mbm{0}_3^{\top} & P_i^{\top} & \mbm{0}_3^{\top}\\
\mbm{0}_3^{\top} & \mbm{0}_3^{\top} & P_i^{\top}
\end{bmatrix}.
\end{array}
\]
\end{Theorem}

Theorem~\ref{th:qform-greeks} shows that, after appropriate elimination of the translation, the camera pose estimation problem can be reduced to the analysis of a quadratic form depending solely on the vectorized rotation matrix. In practice, this implies that the properties of the matrix $\Omega$ play a decisive role in the convergence and stability of optimization-based algorithms.

Based on this result, Terzakis \emph{et al.}~\cite{Terzakis2020} developed the optimal \texttt{SQPnP} algorithm for estimating the extrinsic parameters, which is commonly used in camera calibration to obtain an initial estimate of the pose $(R,t)$ of a calibration object, i.e., to solve the \texttt{PnP} (Perspective-$n$-Point) problem.

In our study, we draw attention to the role of the kernel $\ker(\Omega)$, whose eigenbasis is not explicitly exploited when selecting initial points for the optimization. In the case of a calibration board, i.e., a planar scene where we assume $(P_i)_z=0$ for all points $P_i$ in its local coordinate system, it can be observed that, regardless of the correspondence between scene points and image points $P\mapsto p$, three canonical basis vectors always belong to the kernel, namely $e_3, e_6, e_9\inv{9}\in\ker(\Omega)$. This means that for the third, sixth, and ninth columns of the identity matrix $I_n\eqd[e_1,\dots,e_9]$, the quadratic reconstruction error defined by $\Omega$ is identically zero.

This fact follows directly from the structure of the matrices $\cl{P}_i$, $\cl{T}$, and $\Omega$. Specifically, the explicit form of $\cl{P}_i$ has zero columns at positions $3,6,9$:
\[
P_i \eqd [X_i, Y_i, 0]^\top \lra
\cl{P}_i =
\begin{bmatrix}
X_i & Y_i &  0   & 0   & 0 & 0   & 0 & 0   & 0 \\
0 & 0 & 0 &  X_i & Y_i & 0 & 0   & 0 & 0 \\
0 & 0 & 0 &  0   & 0   & 0 & X_i &  Y_i & 0 
\end{bmatrix}.
\]
Therefore, for any calibration point,
\[
\cl{P}_i e_3 = \cl{P}_i e_6 = \cl{P}_i e_9 = \mbm{0}_3.
\]
Furthermore, the definition of the matrix $\cl{T}$ shows that each term in its defining sum ends with $\cl{P}_i$. Consequently, every term in $\cl{T}$ evaluated on the vectors $e_3, e_6, e_9$ yields zero, and thus both $\cl{T}$ and $\cl{P}_i+\cl{T}$ vanish on these vectors. Since the matrix $\Omega$ is a sum of terms ending with $\cl{P}_i+\cl{T}$, the same reasoning leads to the final conclusion that $\Omega$ annihilates the vectors $e_3, e_6, e_9$.

Summarizing:
\[
\begin{array}{l}
P_i = [X_i, Y_i, 0]^\top \ \lra\
\cl{P}_i e_3=\mbm{0}_3,\ \cl{P}_i e_6=\mbm{0}_3,\ \cl{P}_i e_9=\mbm{0}_3\\
\Lra \cl{T}e_3 = \cl{T}e_6 = \cl{T}e_9 = \mbm{0}_3\\
\Lra \Omega e_3 = \Omega e_6 = \Omega e_9 = \mbm{0}_3.
\end{array}
\]

These properties follow directly from the linear structure of the matrices $\cl{P}_i$ and the definitions of the auxiliary matrices; analogous algebraic arguments are also employed in the analysis of the properties of the Cayley representation (see Appendix~\ref{app:cayley}).

The considerations presented so far have been purely algebraic in nature. In practical calibration tasks, however, the geometry of the experimental setup and the data acquisition process also play an important role. In the next subsection, we discuss the duality between the image plane and the calibration plane, which has direct implications for the interpretation of calibration results and the design of calibration experiments.

\subsection{Duality Between the Image Plane and the Calibration Plane}

The duality between the image plane and the calibration plane arises as a theoretical concept in the context of constructing calibration boards, which define calibration points lying on planar surfaces. The choice of materials on which such points are embedded largely depends on the range of the electromagnetic spectrum in which a given camera operates. In turn, the selected material determines whether the calibration board is mobile while the camera remains fixed, or conversely, whether the board is stationary and the camera is moved.

If a calibration board printed on a sheet of paper is used, the camera is typically fixed, and the board changes its orientation with respect to the camera across successive images. In contrast, when the calibration experiment requires displaying the board on a projector screen, monitor, or television display, successive calibration images must be acquired by changing the pose of the camera.

This naturally leads to the question of whether camera calibration using a board displayed on a screen is equivalent to calibration using a printed board. A related question is what it means for two calibration procedures to be considered equivalent.

We adopt a natural criterion: a small difference in projection errors obtained \emph{from the same set of images} will be taken as evidence of equivalence between the calibration models.

This, however, raises a new issue: can we ensure the same set of identical images in both acquisition modes? Below, we argue that a simple exchange of the camera and board poses, followed by the application of an appropriately chosen isometry -- consisting of a rotation and a translation -- yields the same image.

What does a simple exchange mean in this context? Suppose that, in some reference coordinate system, the calibration board has pose $(U,u)$ and the camera has pose $(V,v)$. Since the camera observes the board, and the board has its $z$-axis aligned with the viewing direction, exchanging their roles requires not only swapping positions but also reversing the directions of the $z$- and $x$-axes. The reversal of the $x$-axis is necessary to preserve a right-handed coordinate system. This transformation is achieved by multiplying the pose components by the orthonormal matrix
\[
E_{xz} =
\begin{bmatrix}
-1 & 0 & 0\\
0 & 1 & 0\\
0 & 0 & -1
\end{bmatrix}
\quad \Lra \quad
U' \eqd E_{xz}U,\;
u' \eqd E_{xz}u,\;
V' \eqd E_{xz}V,\;
v' \eqd E_{xz}v.
\]
After this operation, the camera pose becomes $(U',u')$, while the board pose becomes $(V',v')$.

\begin{figure}[H]
\hspace*{-7mm}
\begin{tabular}{c|c}
\includegraphics[width=0.52\linewidth]{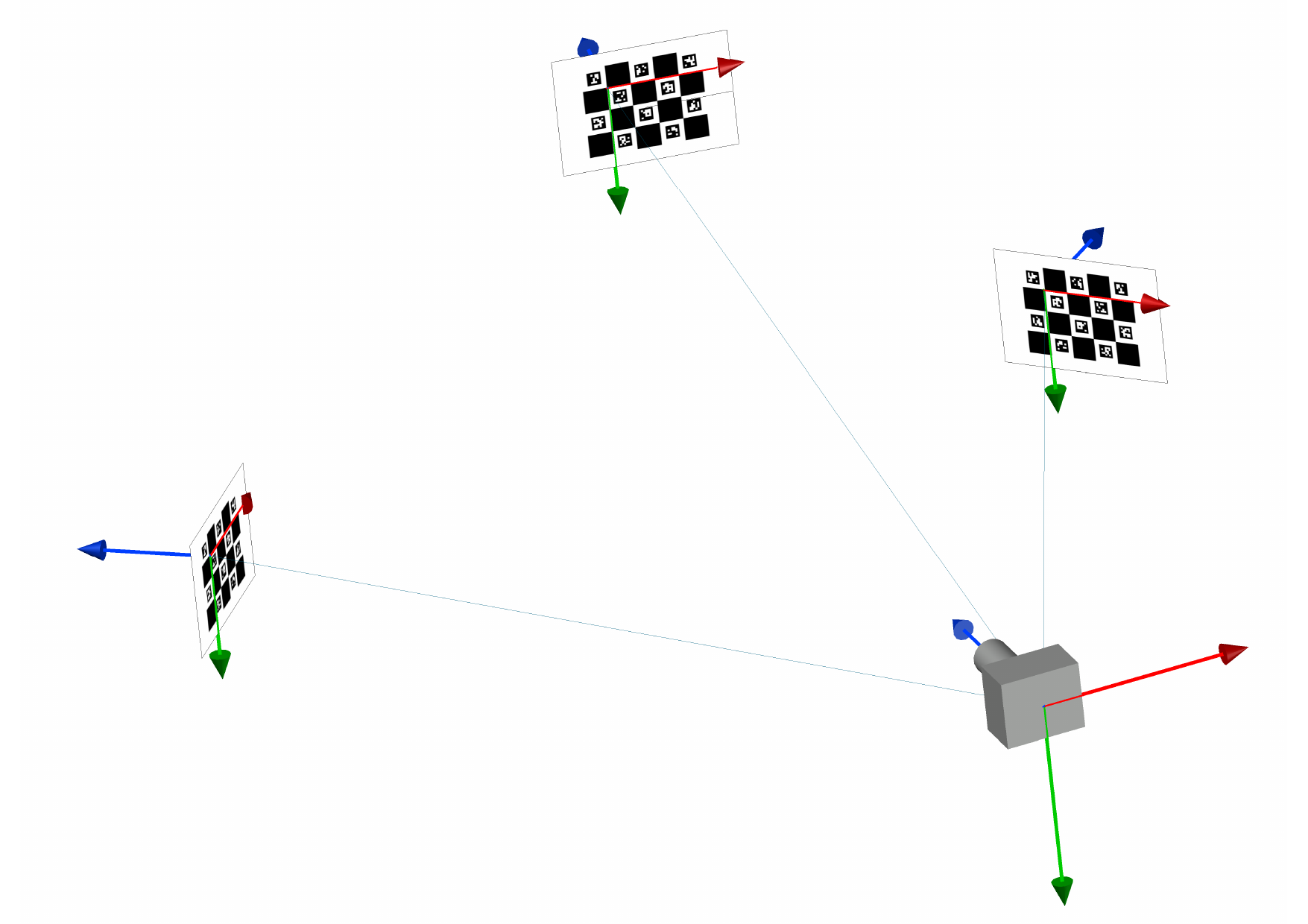}
&
\includegraphics[width=0.52\linewidth]{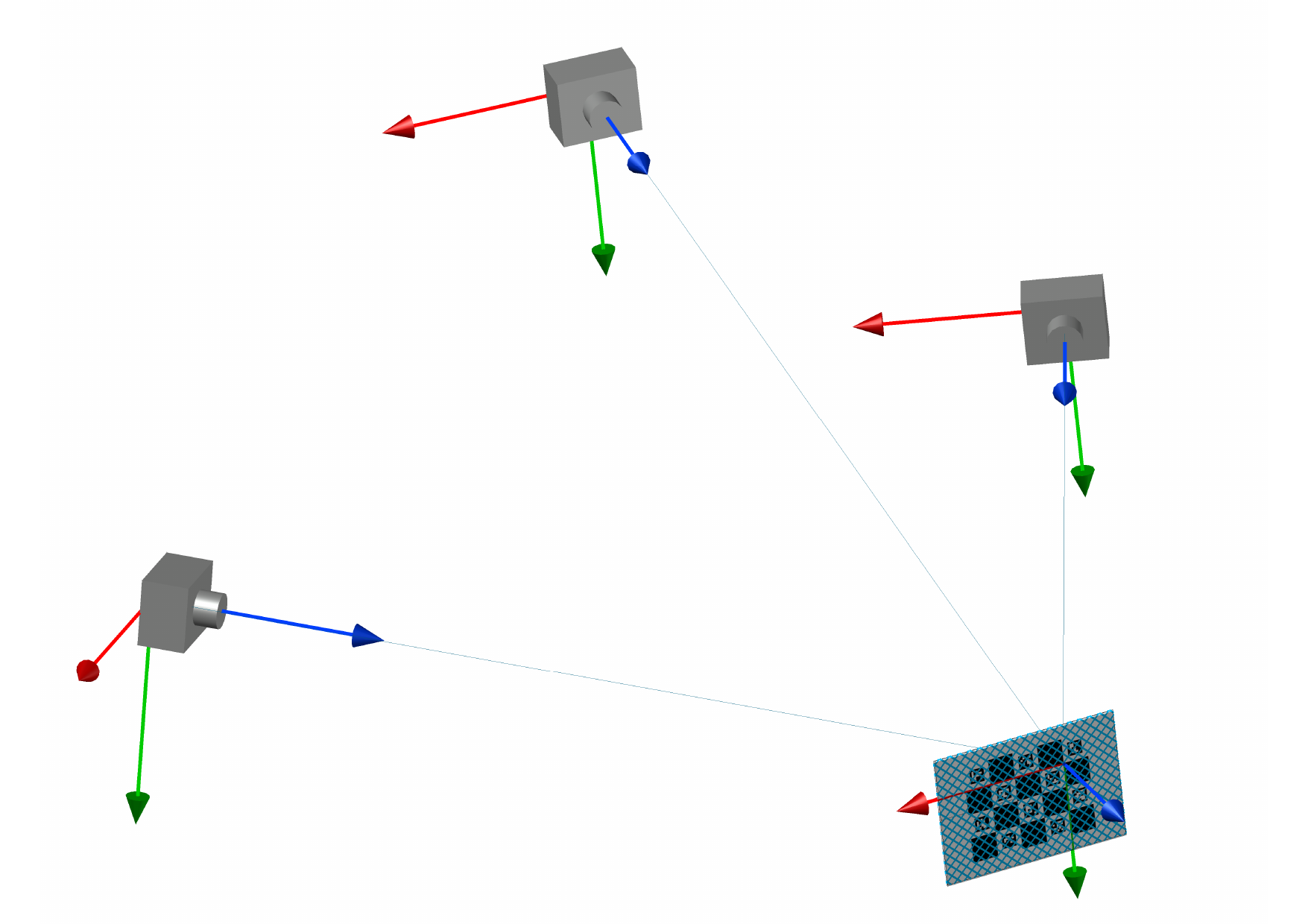}
\end{tabular}
\caption{Dual camera and calibration board configurations: on the left, the camera is stationary and the board is moving; on the right, the board is stationary and the camera is moving.}
\label{fig:du:}
\end{figure}

\begin{Theorem}[On corrective isometry under exchange of camera and board poses]
The corrective isometry is defined by the rotation matrix $W=(R_{vu})^2$ and the translation vector $w=(I_3-R_{vu})t_{vu}$, where the isometry $(R_{vu}, t_{vu})$ represents the transformation from the board coordinate system to the camera coordinate system. Under this transformation, the original image is identical to the image obtained after exchanging the camera and board poses and applying the corrective isometry.
\end{Theorem}

\begin{proof}
Let a point $P$ have coordinates $x$ in the board coordinate system $(U,u)$. Its coordinates in the reference frame are then $y=Ux+u$, and consequently its coordinates in the camera coordinate system $(V,v)$ are given by
\[
y_v=\tp{V}(y-v).
\]
Substituting $y$, we obtain
\[
y_v=\tp{V}(Ux+u-v)
= \underbrace{\tp{V}U}_{\eqd\,R_{vu}}x
+ \underbrace{\tp{V}(u-v)}_{\eqd\,t_{vu}}
= R_{vu}x+t_{vu}.
\]

After exchanging the poses and applying the sign changes to the $x$- and $z$-axes, we use the same expression to compute the transformation from the new board coordinate system $(V',v')$, where $V'\eqd E_{xz}V$ and $v'\eqd E_{xz}v$, to the new camera coordinate system $(U',u')$, where $U'\eqd E_{xz}U$ and $u'\eqd E_{xz}u$:
\[
y_u=\tp{(U')}(V'x+v'-u')
= \tp{U}\underbrace{\tp{E}_{xz}E_{xz}}_{=\,I_3}Vx
+ \tp{U}\underbrace{\tp{E}_{xz}E_{xz}}_{=\,I_3}(v-u)
= \tp{U}Vx+\tp{U}(v-u).
\]

Since, in general, $y_u\neq y_v$, the same image is not obtained directly. We therefore seek an isometry such that $y_v=Wy_u+w$ for all $x\in\mathbb{R}^3$:
\[
(\tp{V}U)x+\tp{V}(v-u)
= W(\tp{U}Vx+\tp{U}(u-v))+w.
\]
From this equation we obtain
\[
W\tp{U}V=\tp{V}U \quad \Lra \quad W=(\tp{V}U)^2=R_{vu}^2,
\]
and
\[
w=\tp{V}(u-v)-W\tp{U}(u-v)
\quad \Lra \quad
w=t_{vu}-\underbrace{W\tp{U}V}_{=\,\tp{V}U\eqd R_{vu}}t_{vu}
=(I_3-R_{vu})t_{vu}.
\]
\end{proof}

The presented duality does not introduce a new calibration model but rather illustrates the geometric equivalence of different experimental configurations. This observation is particularly relevant in the context of calibration using emissive displays such as OLED panels.

In the remainder of the paper, the camera pose estimation problem is addressed using a rotation parameterization in Cayley space. For completeness and notational clarity, we briefly recall the definition of the Cayley rotation matrix and its properties, which are used both in the theoretical analysis and in the description of the \texttt{PnP-ProCay78} algorithm. A more complete, yet concise, summary of the Cayley representation is provided in Appendix~\ref{app:cayley}.

\subsection{Cayley Representation of Rotations}

For completeness and notational clarity, we recall the definition of the Cayley rotation matrix and its basic properties, restricting the presentation to the forms used directly in the remainder of the paper. A more detailed discussion of the Cayley representation can be found in Appendix~\ref{app:cayley}.

\emph{Matrix form}

Let $v\in\mathbb{R}^3$. The Cayley rotation matrix $R(v)\doteq R^{+1}(v)$ and its inverse $R^{-1}(v)=R^\top(v)=R(-v)$ share the following compact matrix representation, using the paired signs $\pm,\mp$:
\[
R^{\pm}
=\mbm{I}_3
\pm\frac{2\cdot\hat{v}\cdot(\mbm{I}_3\pm\hat{v})}{1+\|v\|^2},
\]
where $\hat{v}$ denotes the skew-symmetric matrix associated with the vector $v$, satisfying the standard properties of such matrices (see Appendix~\ref{app:cayley}).

\emph{Element-wise form}

Let $v\in\mathbb{R}^3$. The Cayley rotation matrix $R(v)\doteq R^{+1}(v)$ and its inverse $R^{-1}(v)=R^\top(v)=R(-v)$ admit the following explicit representation, again using the paired signs $\pm,\mp$:
\[
R^{\pm}
=\frac{1}{1+v_x^2+v_y^2+v_z^2}
\begin{bmatrix}
1+v_x^2-v_y^2-v_z^2 & 2(v_xv_y\mp v_z) & 2(v_xv_z\pm v_y)\\
2(v_xv_y\pm v_z) & 1+v_y^2-v_x^2-v_z^2 & 2(v_yv_z\mp v_x)\\
2(v_xv_z\mp v_y) & 2(v_yv_z\pm v_x) & 1+v_z^2-v_x^2-v_y^2
\end{bmatrix}.
\]
The paired signs $\pm$ and $\mp$ correspond respectively to the rotation matrix and its inverse. This property and its geometric consequences are discussed in more detail in Appendix~\ref{app:cayley}.

\section{Initial Estimation of the Calibration Board Pose}

\subsection{Experimental Data}

In this subsection, we describe the experimental dataset used for the initial identification of the calibration board position and orientation with respect to the camera, as well as for evaluating the behavior of the considered PnP algorithms. Particular emphasis is placed on the heterogeneity of the data, encompassing both high-resolution RGB vision cameras and thermal cameras with resolutions on the order of several tens of pixels. Such a configuration allows not only for assessing the accuracy of the methods under strongly varying data quality conditions, but also for analyzing their stability and convergence properties in the context of sensors with extremely different geometric and photometric characteristics.

\begin{figure}[H]
\begin{tabular}{c}
\includegraphics[width=0.9\linewidth]{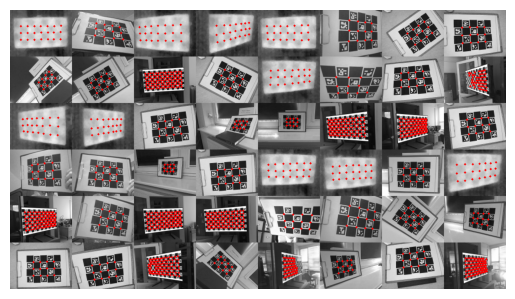}
\end{tabular}
\caption{
Mosaic of representative calibration images acquired using five heterogeneous vision and thermovision cameras, two types of boards with chessboard and multiresolution Charuco grids, and both emissive (OLED screen) and reflective light propagation principles (printed calibration board).
}
\label{fig:mosaic}
\end{figure}

To illustrate the diversity and heterogeneity of the calibration data used in this study, a representative mosaic of calibration images is shown in Fig.~\ref{fig:mosaic}. The dataset comprises images acquired using multiple camera modalities, including high-resolution RGB sensors and low-resolution thermal cameras, resulting in a wide range of spatial resolutions, noise characteristics, and photometric properties.

Each tile in the mosaic corresponds to an individual calibration image, uniformly rescaled to a common display resolution for visualization purposes. Prior to geometric rescaling, all images undergo an achromatic preprocessing step, in which per-pixel chromatic information is suppressed by projecting RGB values onto a single luminance channel. Specifically, the maximum value across color channels is replicated to all channels, yielding a strictly achromatic representation. This operation eliminates chromatic interpolation artifacts that may arise during resolution reduction, particularly for images acquired from OLED displays under low-light conditions.

Detected corner locations are overlaid \emph{after} the mosaic assembly in a common coordinate frame, ensuring resolution-independent visualization and preventing the disappearance of corner markers in low-resolution imagery. This separation of geometric rescaling and feature overlay guarantees consistent marker visibility across different sensor modalities.

An optional random permutation of tiles is employed solely for visualization purposes, avoiding any implicit ordering bias while preserving full reproducibility.

The dataset prepared in this manner serves as a common basis for all subsequent comparative experiments. The same calibration images are used both for analyzing the properties of the matrix~$\Omega$, for selecting the initial points of the \texttt{PnP-ProCay78} algorithm, and for comparisons with reference algorithms available in the \texttt{OpenCV} library.

\begin{figure}[H]
\hspace*{-2mm}
\begin{tabular}{cc}
\includegraphics[width=0.54\linewidth]{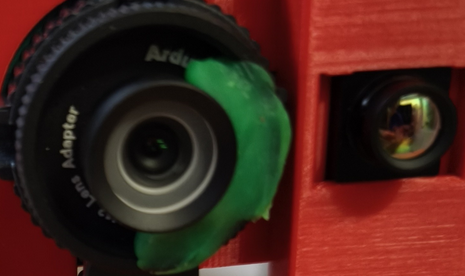}
\includegraphics[width=0.46\linewidth]{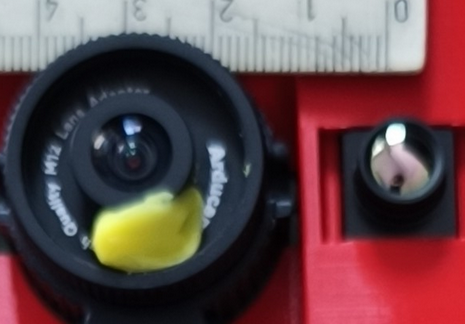}
\end{tabular}
\caption{Views of the RGB and IR cameras used in our experiments.}
\label{fig:rgb-ir-setup}
\end{figure}

Figure~\ref{fig:rgb-ir-setup} shows the physical configuration of the RGB and IR camera setup used in the experiments. On the left, the complete set of sensors mounted on the RP5 (Raspberry Pi~5) computing platform is presented, comprising a high-resolution RGB camera and a low-resolution thermal camera. On the right, a close-up view of the camera lenses is shown, with a millimeter-scale ruler visible in the background. This enables a direct assessment of the distance between the optical centers of the two cameras, which in the considered setup is approximately $20\,\mathrm{mm}$. The small geometric baseline between the cameras facilitates a synchronous analysis of optimization trajectories for RGB and IR data, while preserving comparable observation geometry.

The experimental dataset defined in this way, comprising both high-quality RGB images and spatially highly degraded thermal images, constitutes a challenging test environment for camera pose estimation algorithms. In particular, it enables an assessment of how the algebraic properties of the matrix~$\Omega$ and the choice of initialization points influence the stability and convergence of optimization procedures under conditions of significant differences in image resolution and point localization accuracy. In the following subsection, the described data are used to introduce and discuss in detail the \texttt{PnP-ProCay78} algorithm, whose construction was directly inspired by observations made on this experimental dataset.

\subsection{The \texttt{PnP-ProCay78} Algorithm}

The stage of initial estimation of the calibration object pose, that is, its position and orientation with respect to the camera coordinate system, is crucial not only for accurate pose identification but also for the subsequent estimation of camera parameters, including the distortion model and the sensor plane geometry.

Within the \texttt{OpenCV} framework, several methods for initial pose estimation are provided. Among them, the technique based on the \texttt{SQPnP} algorithm proposed by Terzakis et al.~\cite{Terzakis2020} is distinguished by its direct minimization of the quadratic reconstruction error
$\mathcal{E}^2 \doteq r^\top \Omega r$ (see Definition~\ref{def-recon-error}).  
The algorithm searches for an optimal solution on the sphere $\mathbb{S}(\mathbf{0}_9;\sqrt{3})$, starting from selected saddle points. These points are first projected onto the nearest points on the sphere representing valid rotations, after which a constrained quadratic programming procedure is applied. In a small number of iterations, this process yields a candidate for an optimal rotation.

A candidate saddle point is discarded if the reconstruction error of its nearest rotation exceeds the smallest error found so far. Importantly, saddle points are located in regions of local convexity of the cost function under rotational constraints, which guarantees the existence of a local minimum in such regions, although not necessarily the global one. Consequently, these regions must be systematically explored.  
For planar scenes, the kernel of the matrix $\Omega$ has dimension four, and each nonzero eigenvalue of $\Omega$ defines two symmetric saddle points on the sphere centered at the origin. This leads to at most $2\cdot(9-4)=10$ candidate regions to be examined.

Although \texttt{SQPnP} is currently regarded as one of the most effective solvers for the PnP problem, its construction does not explicitly exploit information arising from the special structure of calibration scenes. In particular, for planar scenes, the properties of the matrix $\Omega$ suggest the possibility of a more directed and deterministic selection of starting points.

Our attempt at \emph{refining an already optimal solution} is therefore based on the following observations regarding the \texttt{SQPnP} technique:
\begin{itemize}
\item The algorithm is designed for both 2D and 3D calibration scenes and thus cannot exploit the specific structure of the quadratic form $r^\top \Omega r$ induced by the assumption $Z_i=0$ for points expressed in the local coordinate system of a planar calibration board. In particular, for planar scenes, the canonical basis vectors $e_3, e_6, e_9$ belong to the kernel of $\Omega$ (see Theorem~\ref{th:qform-greeks}).
\item The reconstruction error values at saddle points, equal to the eigenvalues of $\Omega$ (see Tables~\ref{tab:rgb-omega-eigenvalues} and~\ref{tab:ir-omega-eigenvalues}), are either zero ($\lambda_1=\dots=\lambda_4=0$) or equal to one ($\lambda_7=\dots=\lambda_9$) in the planar case. The eigenvalues $\lambda_5$ and $\lambda_6$ lie very close to the kernel, suggesting that the search should be initiated precisely from the corresponding directions. This behavior, however, is not reflected in the \texttt{SQPnP} pseudocode.
\item The optimization iterations have a Newton-like character but are performed in an extended state vector of dimension $n=9+6=15$. In contrast, the Cayley representation reduces the number of independent parameters to $n=3$. Although the Hessian matrix is approximated by the outer product of the Jacobian, which depends on the number of calibration points, this does not significantly increase the computational cost, especially for low-resolution images such as those encountered in thermal imaging.
\end{itemize}

These observations motivated the following strategy:
\begin{enumerate}
\item Optimization is performed in the Cayley representation of rotations using a generic nonlinear least-squares solver, either \texttt{Trust Region Reflective (TRF)} or \texttt{Levenberg-Marquardt (LM)} \cite{Nocedal2006}.

\item The initial iteration point is selected based on the smallest positive value of the cost function
$r^\top \Omega r = \Omega_{ii}$ evaluated at the canonical basis vectors $[e_1,\dots,e_9] \equiv I_9$.  
Specifically, the vector $\sqrt{3}\,e_{i_{\min}}$ determines the initial Cayley vector $v_1^{(\mathrm{start})}$.
\item The second initial vector is chosen as the antipodal Cayley vector corresponding to the opposite rotation:
$v_2^{(\mathrm{start})} \doteq -v_1^{(\mathrm{start})}$.
\item The solution of the \texttt{lsq} optimization yielding the smaller reconstruction error is selected as the final \texttt{PnP} solution.
\end{enumerate}

This strategy is not heuristic but follows directly from the observed spectral properties of the matrix $\Omega$ and their consistency across a wide range of experimental data.

In the following subsections, we demonstrate the consequences of selecting $v_{\mathrm{start}}$ based on canonical basis vectors and present the final form of the algorithm, which we name \texttt{PnP-ProCay78} after \emph{Procrustes}, \emph{Cayley}, and the indices $7$ and $8$ of the selected canonical vectors.

\subsubsection{How Are the Initial Cayley Vectors Selected?}

To determine which canonical vectors should be chosen, we analyze the frequency statistics of minimal-cost occurrences and the average cost values associated with individual canonical directions (Table~\ref{tab:repro-stats}). These statistics are computed from several dozen calibration images acquired during the calibration of five different cameras. Representative raw data used to compute these statistics are provided in Appendix~\ref{app:canonical-error-values} for experiments involving RGB and IR cameras mounted on the RP5 (Raspberry Pi~5) computing platform.

\begin{table}[H]
\centering
\caption{Histogram of canonical vector indices corresponding to the minimum positive reconstruction error.The mean error value is also reported for each canonical vector group.}
\label{tab:repro-stats}
\small
\begin{tabular}{lcccc}
\toprule
camera name & \#photos & histogram at minimum (\%)& error means for canonical groups\\
\midrule
RP5-RGB (2048x1520) & $31$ & $0_{1-6},\,42_7,\,58_8,\,0_9$
& $0_{3,6,9},\ 0.00049_{8,7},\, 0.51_{5,1},\, 0.99_{4,2}$\\
RP5-IR (80x62)      & $22$ & $0_{1-6},\,32_7,\,68_8,\,0_9$
& $0_{3,6,9},\ 0.0032_{8,7},\, 0.51_{5,1},\, 0.99_{4,2}$ \\
PC-Aver (640X480) 	& 12 & $0_{1-6},\,58_7,\,42_8,\,0_9$
&$0_{3,6,9},\ 0.0057_{7,8},\, 0.53_{1,5},\, 0.97_{4,2}$\\
PC-Logitex (640x480)& 7 & $0_{1-6},\,43_7,\,57_8,\,0_9$
& $0_{3,6,9},\ 0.0035_{8,7},\, 0.58_{5,1},\, 0.92_{4,2}$\\
PC-USB (640x480)    & 8 & $0_{1-6},\,50_7,\,50_8,\,0_9$
& $0_{3,6,9},\ 0.0067_{7,8},\, 0.50_{1,5},\, 0.99_{4,2}$\\
\bottomrule
\end{tabular}
\end{table}

Table~\ref{tab:repro-stats} leads to an unambiguous conclusion: for every calibration image, the smallest positive reconstruction error is attained either for the canonical vector $e_7$ or for $e_8$.

This observation allows the initial Cayley vectors $v_7$ and $v_8$ to be determined independently of the specific input data used by the \texttt{PnP-ProCay78} algorithm.

The computation follows the scheme
\[
\sqrt{3}\,e_i
\;\xrightarrow{\text{vector to matrix}}\;
E_i
\;\xrightarrow{\text{Procrustes}^+}\;
R_i
\;\xrightarrow{\text{Cayley}}\;
v_i .
\]
Details of the Procrustes procedure and its modification used in this work are provided in Appendix~\ref{app:procrustes}.

Table~\ref{tab:v7-v8-v9} lists the results of these computations for $e_7$ and $e_8$. For completeness, the case of $e_9$ is also included. Its nearest rotation is the identity matrix $I_3$, corresponding to a zero Cayley vector. The antipodal matrix $-I_3$ has trace equal to $-1$ and therefore does not admit a Cayley representation.

From Table~\ref{tab:v7-v8-v9}, the two initial vectors required for the two least-squares optimizations are determined as follows:
\begin{itemize}
\item if $e_7^\top \Omega e_7 < e_8^\top \Omega e_8$, the starting Cayley vectors are
$v_7 = [0,-1,0]^\top$ and $\bar v_7 = [0,1,0]^\top$;
\item otherwise, the pair
$v_8 = [1,0,0]^\top$ and $\bar v_8 = [-1,0,0]^\top$ is used.
\end{itemize}
\begin{table}[H]
\centering
\caption{
Cayley representations of the nearest rotation matrices and their antipodes computed for the selected canonical points $\sqrt{3}e_i$ on the search sphere $S(\mbm{0}_9;\sqrt{3})$ in the Euclidean space $\bb{R}^9$, for $i=7,8,9$.}
\label{tab:v7-v8-v9}
\small
\begin{tabular}{cccc|cc}
\toprule
$\sqrt{3}e_i$ & 
matrix $E_i$ & 
nearest $R_i$ &  
Cayley $v_i$ & 
antipodal $\ov{R}_i$ & 
Cayley $\ov{v}_i$
\\
\midrule
$\sqrt{3}e_7$ 
& 
	$\begin{bmatrix}0&0&0\\0&0&0\\\sqrt{3}&0&0\end{bmatrix}$
&  
	$\begin{bmatrix}0&0&-1\\0&1&0\\1&0&0\end{bmatrix}$ 
& 
	$\begin{bmatrix}0\\-1\\0\end{bmatrix}$
& 
	$\begin{bmatrix}0&0&1\\0&1&0\\-1&0&0\end{bmatrix}$ 
& 
	$\begin{bmatrix}0\\1\\0\end{bmatrix}$  
\\
\midrule
	$\sqrt{3}e_8$ 
& 
	$\begin{bmatrix}0&0&0\\0&0&0\\0&\sqrt{3}&0\end{bmatrix}$
& 
	$\begin{bmatrix}1&0&0\\0&0&-1\\0&1&0\end{bmatrix}$ 
& 
	$\begin{bmatrix}1\\0\\0\end{bmatrix}$ 
& 
	$\begin{bmatrix}1&0&0\\0&0&1\\0&-1&0\end{bmatrix}$ 
& 
	$\begin{bmatrix}-1\\0\\0\end{bmatrix}$  
\\
\midrule
\midrule
$\sqrt{3}e_9$
& 
	$\begin{bmatrix}0&0&0\\0&0&0\\0&0&\sqrt{3}\end{bmatrix}$
& 
	$\begin{bmatrix}1&0&0\\0&1&0\\0&0&1\end{bmatrix}$ 
& 
	$\begin{bmatrix}0\\0\\0\end{bmatrix}$ 
& 
	$\begin{bmatrix}1&0&0\\0&-1&0\\0&0&-1\end{bmatrix}$ 
& 
$\begin{array}{c}Undefined\ as\\\tr{\ov{R}}=-1 \end{array}$
\\
\bottomrule
\end{tabular}
\end{table}

Two independent optimization runs starting from antipodal regions of the search sphere are necessary due to the symmetry of the cost function:
$r^\top \Omega r = (-r)^\top \Omega (-r)$.
Starting from $R_7$, for example, the opposite point $-R_7$ yields the same cost value. Since negating a rotation matrix changes its determinant to $-1$, an Umeyama correction~\cite{Umeyama1991} is required:
\[
\begin{array}{l}
E_i = \mathrm{mat}(\sqrt{3}e_i)
\xrightarrow{\text{Procrustes}}
R = [u_1,u_2,u_3]V^\top
\xrightarrow{\text{point symmetry}}\\[4pt]
-R = [-u_1,-u_2,-u_3]V^\top
\xrightarrow{\text{Umeyama correction}}
\bar R = [-u_1,-u_2,u_3]V^\top .
\end{array}
\]
This extension of the Procrustes procedure with the Umeyama correction is referred to as Procrustes$^+$.  
A formal proof of its correctness and an analysis of solution uniqueness are provided in Appendix~\ref{app:procrustes}.

Finally, we note the following:
\begin{enumerate}
\item The Umeyama correction is typically applied when the Procrustes solution $R=UV^\top$ has a negative determinant. In our case, all matrices $R_i=\mathrm{Procrustes}(\mathrm{mat}(\sqrt{3}e_i))$, $i=1,\dots,9$, have determinant $+1$, whereas their antipodes have determinant $-1$, and the correction is applied only to the latter.
\item In general, an antipodal rotation does not correspond to the negation of the Cayley vector, since $R_{-v} = R_v^\top \neq -R_v$. However, for the specific cases of $R_7$ and $R_8$ listed in Table~\ref{tab:v7-v8-v9}, this equality holds.
\item The starting points in \texttt{SQPnP} are scaled eigenvectors of $\Omega$ and thus depend on all matrix entries. In contrast, the starting points of \texttt{PnP-ProCay78} depend solely on the comparison of the diagonal elements $\Omega_{77}$ and $\Omega_{88}$:
\[
\begin{array}{l}
\Omega_{77}<\Omega_{88}
\;\Longrightarrow\;
\left\{
\begin{bmatrix}0\\-1\\0\end{bmatrix},
\begin{bmatrix}0\\1\\0\end{bmatrix}
\right\},\\[24pt]
\Omega_{77}\geq\Omega_{88}
\;\Longrightarrow\;
\left\{
\begin{bmatrix}1\\0\\0\end{bmatrix},
\begin{bmatrix}-1\\0\\0\end{bmatrix}
\right\}.
\end{array}
\]
\end{enumerate}

\subsubsection{The \texttt{PnP-ProCay78} Algorithm: Pseudocode}

\begin{table}[H]
\caption{The \texttt{PnP-ProCay78} algorithm and the \texttt{ReconstructionResidual} function (required by the \texttt{LSM} optimizer).}
\label{alg:procay}
\centering
\begin{tabular}{@{}p{0.50\linewidth} p{0.46\linewidth}@{}}

\textbf{Algorithm 1: PnP-ProCay78}

\begin{algorithmic}[1]
\Require $\left\{\begin{array}{l}
 P \in \mathbb{R}^{n \times 3} \text{ -- board points with } P_{:,z}=\mathbf{0}_n,\\
 p \in \mathbb{R}^{n \times 3} \text{ -- image points with } p_{:,z}=\mathbf{1}_n,\\
 \Omega \text{ -- reconstruction error matrix},\\ 
 \mathcal{T} \text{ -- translation operator (computing } t \text{ from } R)
\end{array}\right.$

\Ensure $(R,t)$ with minimal reconstruction error

\State $v_{\text{start}} \gets
\begin{cases}
[0,-1,0]^\top, & \textbf{if } \Omega_{77} < \Omega_{88},\\
[1,0,0]^\top, & \textbf{otherwise}.
\end{cases}$

\State $(v,\gamma) \gets
\textsc{LSM}_{trf}(\textsc{RR}, v_{\text{start}}; P, p, \mathcal{T})$

\State $(\bar v,\bar\gamma) \gets
\textsc{LSM}_{trf}(\textsc{RR}, -v_{\text{start}}; P, p, \mathcal{T})$
\Statex (* \textsc{LSM}$_{trf}$ denotes the \texttt{trf}-based least-squares solver returning a Cayley vector and its associated loss value *)

\State \textbf{if } $\bar\gamma > \gamma$ \textbf{ then } 
$v^\star \gets v$ \textbf{ else } $v^\star \gets \bar{v}$

\State $R \gets \textsc{CayleyToMatrix}(v^\star)$
\State $t \gets \mathcal{T}\,\mathrm{vec}(R)$

\State \Return $(R,t)$
\end{algorithmic}

&
\textbf{Algorithm 2: \textsc{RR} (Reconstruction Residuals)}

\begin{algorithmic}[1]
\Require $\left\{\begin{array}{l}
v \text{ -- current Cayley vector},\\
P, p, \mathcal{T} \text{ -- as in \texttt{PnP-ProCay78}}
\end{array}\right.$
\Ensure residual error vector $e_{\text{vec}}$

\State $R \gets \textsc{CayleyToMatrix}(v)$
\State $t \gets \mathcal{T}\,\mathrm{vec}(R)$
\State $Q \gets P R^{\top} + t$

\State $E \gets (Q \oslash Q_{:,z} - p \odot Q_{:,z})_{:,xy}$
\Statex \textit{(* Elementwise operations are broadcasted *)}
\State $e_{\text{vec}} \gets \mathrm{vec}(E)$
\State \Return $e_{\text{vec}}$
\end{algorithmic}

\end{tabular}
\end{table}

It is worth emphasizing that the proposed \texttt{PnP-ProCay78} algorithm employs a hybrid use of cost functions that play complementary roles in the estimation process.
Like in \texttt{SQPnP} algorithm, the translation vector $t$ is eliminated analytically by enforcing the stationarity condition of the \emph{reconstruction error} with respect to $t$, which leads to a closed-form linear operator $\mathcal{T}$ mapping the rotation matrix $R$ to the corresponding optimal translation.
In contrast, the nonlinear optimization stage (TRF or LM) operates directly on \emph{projection residuals}, which constitute the primary application-level metric of pose accuracy.

Although the final optimization stage minimizes projection residuals, the translation component is obtained analytically by enforcing stationarity of the quadratic reconstruction error with respect to translation. This hybrid design separates metric alignment from optimization efficiency.

This separation mirrors a common design pattern observed in modern optimization and learning-based methods \cite{Bottou2018}, where a surrogate objective is used to structure the model efficiently, while the final optimization is driven by the target performance metric.
In this sense, \texttt{PnP-ProCay78} can be interpreted as a geometrically motivated hybrid approach that combines an analytically tractable surrogate (reconstruction error) with a physically meaningful loss function (projection error).

\subsubsection{\texttt{PnP-ProCay78} versus \texttt{PnP-OpenCV}: Experimental Results}

In our experiments, we compared the camera pose identification results of our \texttt{PnP-ProCay78} algorithm with three leading \texttt{PnP} algorithms included in the \texttt{OpenCV} library: \texttt{SQPnP, IPPE, EPnP}. Measurements were performed for five different cameras. Two of them, marked in the table as RP5-RGB and RP5-IR, were integrated into a Raspberry Pi 5 system. Two other cameras, marked here as PC-Logit and PC-USB, are inexpensive webcams connected to PCs. The fifth camera is a slightly more expensive laboratory system used primarily by lecturers for remote handwriting image transmission.

\begin{table}[H]
\centering
\caption{
Comparison of error statistics (RMSE, median, max) for four PnP methods and five cameras.
For each camera, in addition to the projection error computed in the camera projective plane, the reprojection error in pixels is also estimated. For this purpose, a preliminary estimation of the focal length in pixels is used (see column $f_{\text{pixel}}$).
All values in the table are normalized by the coefficient $scale$ (see column $scale$), i.e.,
$true\_value = value \cdot scale$. For \texttt{PnP-ProCay78} algorithm the \texttt{TRF} was used. However, for \texttt{LM} optimizer the results are nearly the same.
}
\label{tab:repro-stats}
\small
\begin{tabular}{l|c|cccc}
\toprule
Camera & & ProCay78 & SQPnP \cite{Terzakis2020} & IPPE \cite{Collins2014} & EPnP \cite{Lepetit2009} \\
\midrule
$(f_{\text{pixel}})$ & scale & (rmse, med, max) & (rmse, med, max) & (rmse, med, max) & (rmse, med, max) \\
\midrule
RP5-RGB & $10^{-4}$      & 
$(0.35,\, 0.24,\, 1.03)$ & 
$(0.35,\, 0.24,\, 1.02)$ & 
$(0.35,\, 0.24,\, 1.03)$ & 
$(10.33,\, 3.38,\, 68.21)$ 
\\
$(2175)$ & $10^{-1}$       & 
$(0.76,\, 0.51,\, 2.24)$   & 
$(0.76,\, 0.51,\, 2.22)$   &
$(0.76,\, 0.52,\, 2.23)$   & 
$(22.46,\, 7.35,\, 148.33) $
\\
\midrule
RP5-IR & $10^{-4}$          & 
$(10.90,\, 6.67,\, 28.27)$  & 
$(10.92,\, 6.41,\, 29.13)$  & 
$(11.08,\, 6.50,\, 29.14)$  & 
$(111.55,\, 53.20,\, 347.48)$ 
\\
$(97)$ & $10^{-1}$        & 
$(1.05,\, 0.64,\, 2.73)$  & 
$(1.05,\, 0.62,\, 2.81)$  & 
$(1.07,\, 0.63,\, 2.81)$  & 
$(10.77,\, 5.14,\, 33.56) $ 
\\
\midrule
PC-Logitex & $10^{-4}$   & 
$(1.18,\, 0.65,\, 3.38)$ & 
$(1.18,\, 0.64,\, 3.38)$ & 
$(1.48,\, 0.70,\, 3.54)$ & 
$(4.73,\, 3.13,\, 13.67) $
\\
$(563)$ & $10^{-1}$      & 
$(0.66,\, 0.37,\, 1.90)$ & 
$(0.66,\, 0.36,\, 1.90)$ & 
$(0.83,\, 0.40,\, 1.99)$ & 
$(2.66,\, 1.76,\, 7.69)  $
\\
\midrule
PC-Aver & $10^{-4}$      & 
$(1.35,\, 0.73,\, 4.69)$ & 
$(1.35,\, 0.72,\, 4.68)$ & 
$(1.55,\, 0.95,\, 4.20)$ & 
$(4.19,\, 1.78,\, 20.94)$
\\
$(604)$ & $10^{-1}$      & 
$(0.81,\, 0.44,\, 2.84)$ & 
$(0.81,\, 0.44,\, 2.83)$ & 
$(0.93,\, 0.58,\, 2.54)$ & 
$(2.53,\, 1.07,\, 12.66) $  
\\
\midrule
PC-USB & $10^{-4}$        & 
$(2.70,\, 2.46,\, 5.70)$  & 
$(2.71,\, 2.49,\, 5.44)$  & 
$(2.78,\, 2.78,\, 5.27)$  & 
$(14.34,\, 8.28,\, 33.80) $
\\
$(633)$ & $10^{-1}$ & 
$(1.71,\, 1.56,\, 3.61)$ & 
$(1.71,\, 1.58,\, 3.44)$ & 
$(1.76,\, 1.76,\, 3.34)$ & 
$(9.08,\, 5.24,\, 21.40) $
\\
\bottomrule
\end{tabular}
\end{table}

Table \ref{tab:repro-stats} contains for each combination (camera, PnP algorithm) the following statistics of projection (first row) and reprojection (second row) errors: rmse (Root Mean Square Error), median absolute error, maximum absolute error.

From the table, we conclude that for each camera, the \texttt{PnP-ProCay78} and \texttt{SQPnP} algorithms have almost the same (re)projection error statistics. The \texttt{IPPE} algorithm has also nearly the same statistics, because the difference in hundredths of a pixel is insignificant. The \texttt{EPnP} algorithm is much worse than these three leading algorithms. 

\subsubsection{Analysis of State Trajectories in Cayley Space}

Visual analysis of optimization trajectories in Cayley space allows direct observation of differences in the behavior of algorithms, as well as the impact of input data quality (RGB vs. IR) on the convergence process.

To this end, any nonzero vector $v\inv{3}, v=\tp{[v_x, v_y, v_z]}$ is represented by a point
$v_{xy}'\eqd(v_x/\|v\|, v_y/\|v\|)$ lying on the unit disk and a real number $v_n\eqd \operatorname{sign}'(v_z)\cdot\|v\|$. The zero vector R$v\eqd\mbm{0}_3$ is represented by the zero vector in the plane and the number zero. 

The sign'$(t)$ function used above is computed according the settings of sign bit in float representation, i.e. sign'$(-0.0)=-1$ and  sign'$(+0.0)=+1$ what in general means that if the sign bit set then sign' returns $-1$, otherwise sign' returns $+1$.

Such a vector decomposition is called its radially circular representation. This representation has several interesting properties:
\begin{enumerate}
\item It is invertible: $\|v\|= |v_n|,\ (v_x,v_y) = v_n\cdot v_{xy}',\ v_z = \operatorname{sign}'(v_n)\cdot\sqrt{\|v\|^2-v_x^2-v_y^2}$.
\item Each point in the plane $v_z=0$ is mapped to a point on the unit circle.
\item Any trajectory in 3D space can be observed as a trajectory in 2D space parallel to the signed norm curve of the trajectory points.
\end{enumerate}

\begin{figure}[H]
\begin{tabular}{cc}
\includegraphics[width=0.48\linewidth]{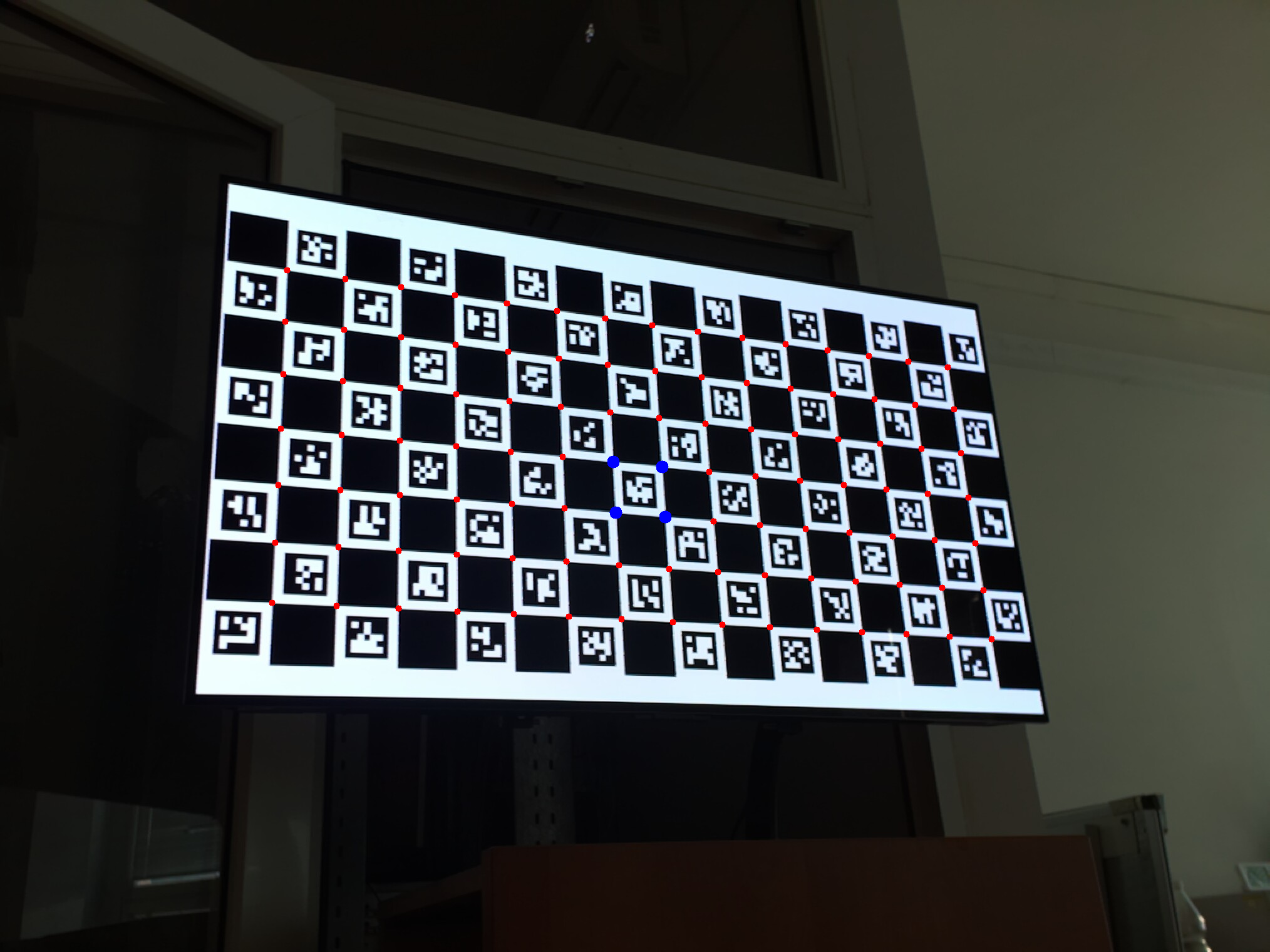}
&
\includegraphics[width=0.48\linewidth]{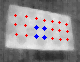}
\\
\includegraphics[width=0.48\linewidth]{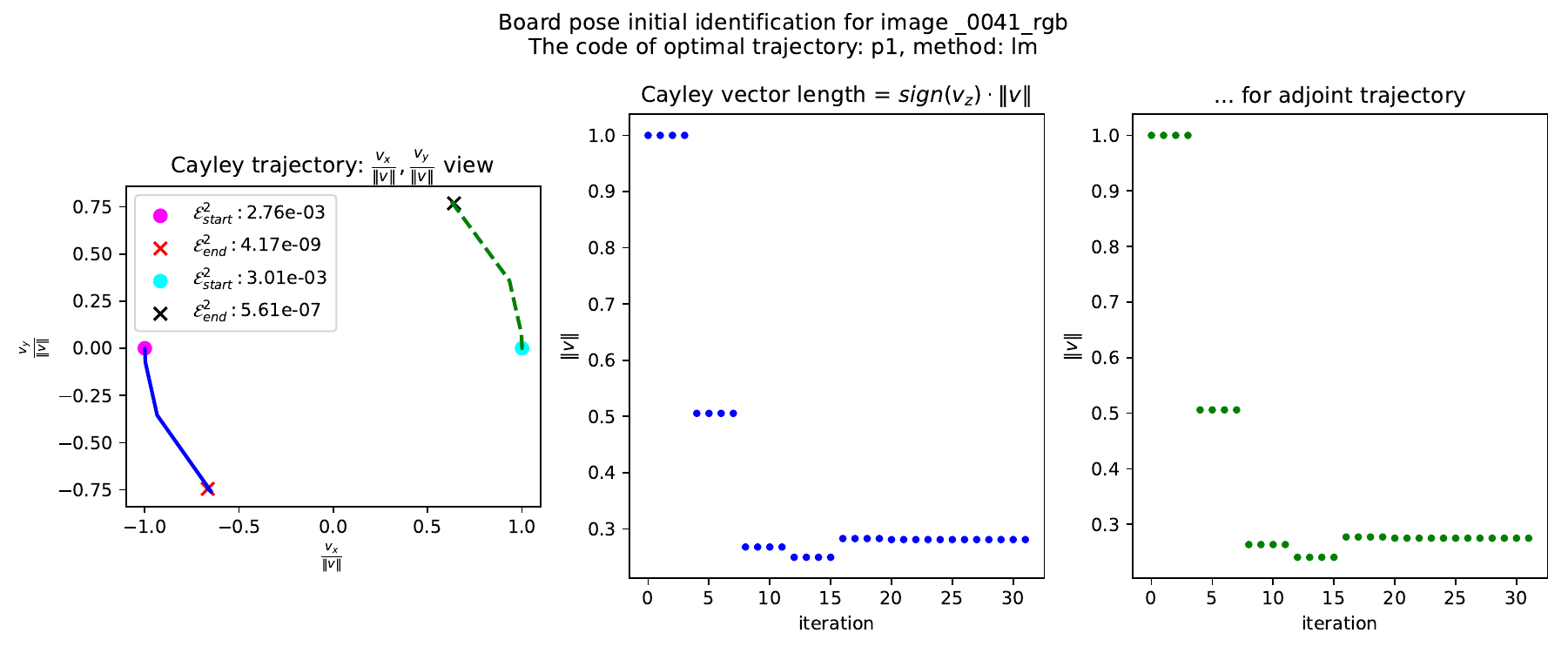}
&
\includegraphics[width=0.48\linewidth]{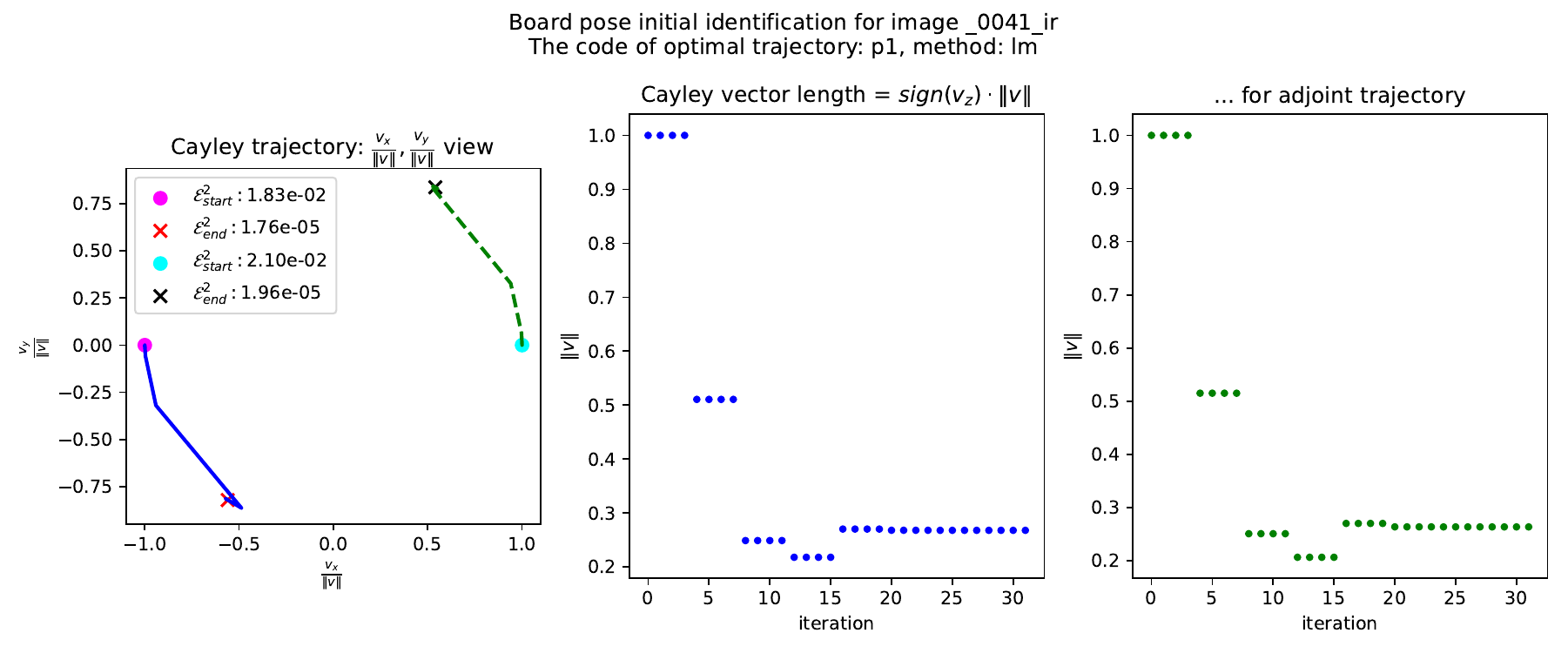}
\end{tabular}
\caption{
Calibration board views (id=41) displayed sequentially on an OLED screen:
on the left, a board with Charuco markers; on the right, a thermal image of a standard chessboard.
The chessboard grid perfectly overlaps the Charuco grid at half the resolution.
In the second row, trajectories of the LM optimizer states in Cayley space are shown
for RGB camera pose estimation (left) and IR camera pose estimation (right), respectively.
}
\label{fig:traj-41}
\end{figure}

Figure \ref{fig:traj-41} shows images of calibration grids displayed on an OLED screen. The image on the left was acquired by an RGB camera displaying a Charuco grid \cite{Garrido2014}, while the image on the right was captured by a low-resolution thermal camera displaying a traditional, lower-resolution checkerboard. There are descriptions of thermal camera calibration using Charuco grids in scientific publications, such as \cite{Roshan2024}, but in our case, detecting calibration points in an 80x62 resolution thermal image using Charuco tiles failed. In high-resolution images, detection of Aruco markers is effective, even under challenging lighting conditions; see \cite{hu2019}.

In the second row of Figure \ref{fig:traj-41}, below the image from the \texttt{RGB} camera, we have depicted Cayley trajectories using a radial-circular visualization. The solid blue line connects the Cayley states for the iterations of the \texttt{LSQ-Levenberg-Marquardt} algorithm, while the dashed green line represents its performance for the conjugate trajectory for states with higher projection error. 

The label (code) of the trajectory indicates whether the initial Cayley vector represented the closest orthonormal matrix with a positive determinant—code \texttt{p0}—or with a negative determinant and the required Umeyama correction—code \texttt{p1}. 

To the right of the trajectory within the unit disk, 
the signed norm values are plotted with stepwise indexing. 
The first step equals to $v_n=1$ as the initial Cayley vectors have norms equal to one, and $z$ coordinates zero, see table \ref{tab:v7-v8-v9}. The next three points are the same  as the specific procedural warm-up of \texttt{Scipy least\_squares.py} functionality. The final series of near-constant state norms leads to the conclusion that a small reduction in projection error can be achieved after step 15. Therefore, more than $50\%$ of iterations could be avoided by observing only the stabilization of the loss function, ignoring gradient energy decay.

In the second row of Figure 41, below the image from the IR camera, we also depicted the Cayley trajectories using a radial-circular visualization for the thermal image of the same view of the board. We can observe a high visual correlation of the trajectories for identifying the positions of the visual and thermal cameras.

In the next Figure 41a, we present both visualizations of the \texttt{Trust Region with Reflection (TRF)}  trajectories at a larger scale – the upper one for the RGB camera and the lower one for the IR camera. More experimental drawings can be found in the Appendix \ref{app:traj-05-20}.

\begin{figure}[H]
\begin{tabular}{c}
\includegraphics[width=0.99\linewidth]{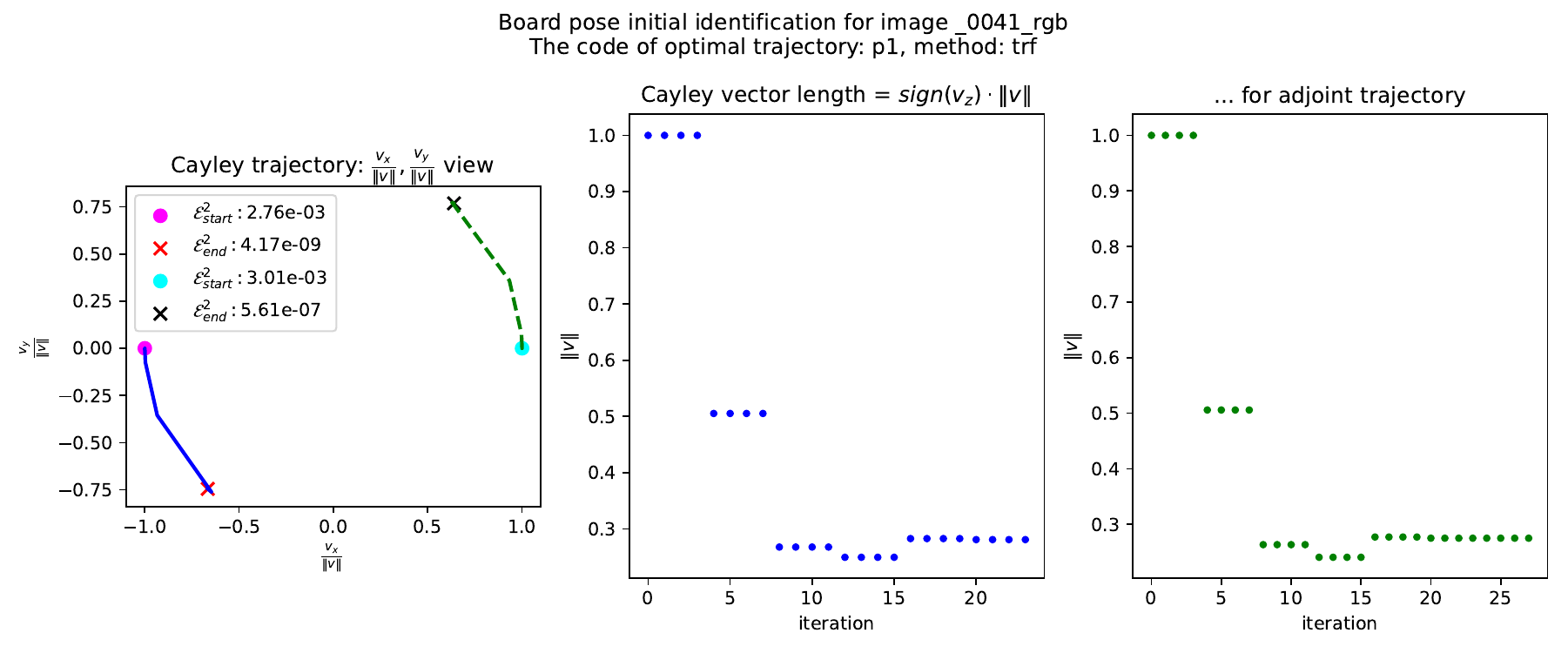}
\\
\includegraphics[width=0.99\linewidth]{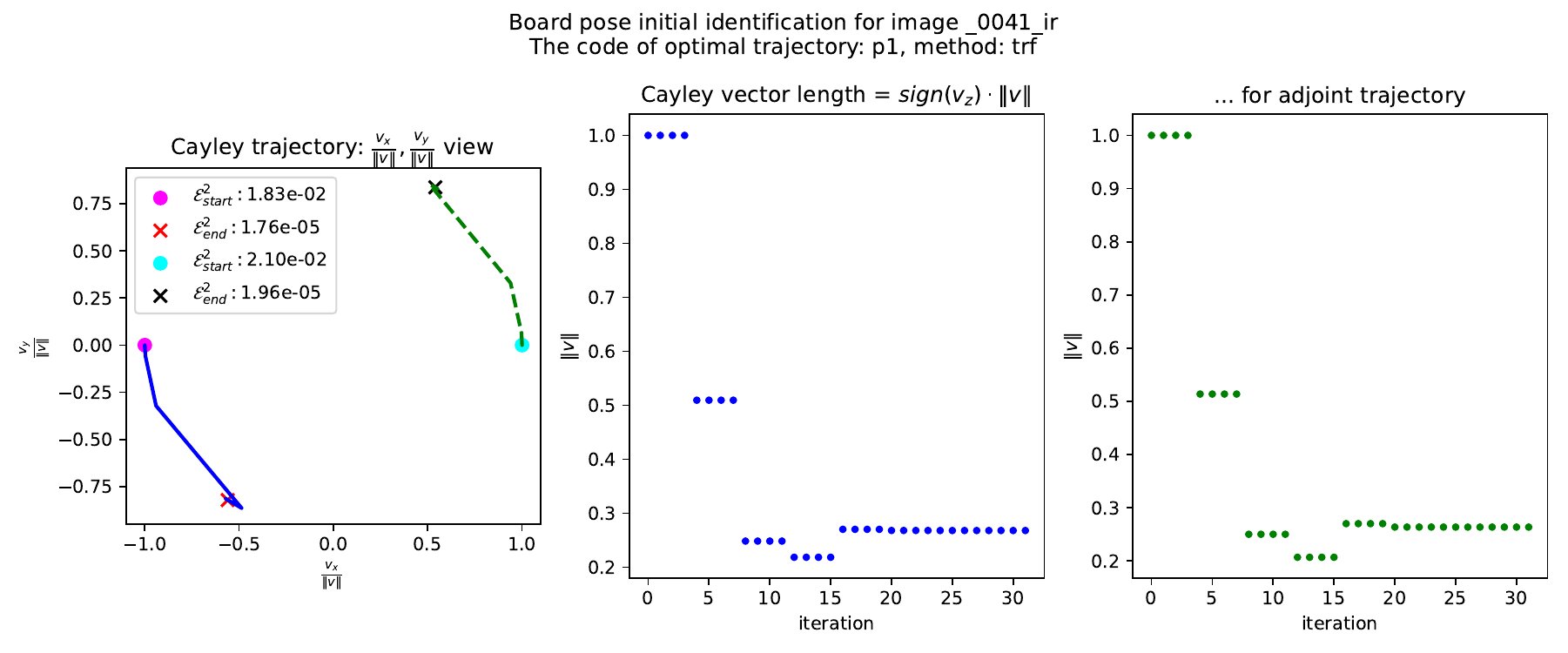}
\end{tabular}
\caption{
Trajectories of the TRF optimizer states in Cayley space.
The first row corresponds to RGB camera pose estimation,
while the second row shows the trajectory for the low-resolution thermal camera.
}
\label{fig:traj-41a}
\end{figure}

\section{Discussion and Conclusions}

In this work, we proposed the \texttt{PnP-ProCay78} algorithm as an alternative approach to estimating the camera pose with respect to a planar calibration object. The starting point is the classical quadratic formulation of the reconstruction error, known from the \texttt{SQPnP} algorithm; however, the key differences concern the selection of initial points and the parameterization of rotations.

The most important theoretical observation is the exploitation of the kernel structure of the reconstruction error matrix $\Omega$ in the planar case. It was shown that the canonical basis vectors $e_3$, $e_6$, and $e_9$ always belong to $\ker(\Omega)$, while the minimal positive values of the cost function consistently occur along the directions $e_7$ and $e_8$. Since the value of the quadratic form $\tp{x}\Omega x$ corresponds to the reconstruction error for $x\in\mathbb{R}^9$, it can be computed directly from the definition of the reconstruction error itself. As a result, during the initialization stage, the \texttt{PnP-ProCay78} algorithm does not require the explicit construction of the matrix $\Omega$.

However, in order to continue the optimization process efficiently, the algorithm requires fast computation of the translation vector using the relation $t=\cl{T}r$, where $\cl{T}\in\mathbb{R}^{3\times 9}$ is a constant, data-dependent matrix. For this reason, the algorithm relies on the stationarity condition of the reconstruction error function $\cl{E}^2$ with respect to translation. There is no direct dependence on the numerical values of the reconstruction error itself; instead, the dependence is on the minimizing translation
\[
t_{\min}\eqd \arg\min_{t\in\mathbb{R}^3} \cl{E}^2(r,t)=\cl{T}r.
\]
By using the reprojection error directly as the loss function in the nonlinear optimizer, while indirectly relying on the reconstruction error through the stationarity condition, the \texttt{PnP-ProCay78} algorithm can be regarded as a hybrid method with respect to these two loss functions.

Unlike \texttt{SQPnP}, which initializes the optimization from saddle points obtained via a full spectral decomposition, the \texttt{PnP-ProCay78} algorithm relies solely on the relation between the diagonal elements $\Omega_{77}$ and $\Omega_{88}$. This leads to an unambiguous and fully deterministic selection of two antipodal starting points in Cayley space.

The use of the Cayley parameterization enables a reduction of the state-space dimensionality to three independent variables, allowing standard least-squares solvers such as the Trust Region Reflective method and the Levenberg--Marquardt algorithm to be employed. An analysis of the optimization trajectories in Cayley space provides additional visual insight into the convergence process, both for high-resolution RGB cameras and for very low-resolution thermal cameras. Notably, despite substantial differences in input data quality, the trajectories exhibit strong spatial similarity and close convergence for both modalities.

Comparative experimental results indicate that, in terms of reconstruction error, \texttt{PnP-ProCay78} achieves accuracy practically identical to \texttt{SQPnP} and slightly superior to \texttt{IPPE}, while maintaining a considerably simpler algorithmic structure. In practice, the computation time for complete calibration sequences remains negligible compared to other stages of the calibration pipeline, rendering computational complexity a secondary concern.

An additional contribution of this work is the demonstration of a coherent pose analysis for RGB and thermal cameras within a multi-sensor system. Although the \texttt{PnP-ProCay78} algorithm does not explicitly incorporate sensor-specific information, the experiments show that a joint analysis of optimization trajectories can serve as a valuable diagnostic tool in integrated RGB--IR vision systems.

In summary, the proposed \texttt{PnP-ProCay78} algorithm offers a conceptually simple, geometrically transparent, and didactically attractive alternative to existing PnP solutions for planar scenes. Its reliance on standard geometric models and classical optimization techniques makes it particularly suitable for both research applications and educational use. Potential directions for future work include extending the method to quasi-planar scenes and investigating its properties in the context of joint calibration of multi-camera systems.

\section*{References}
\bibliography{references}

\appendix

\section{Cayley Representation for $\SO{n}$}
\label{app:cayley}

This appendix collects the basic definitions and properties of the Cayley representation \cite{Cayley1846} for the group $\SO{n}$ that are used throughout the paper, in particular in the analysis of rotation parametrization and in the study of the \texttt{PnP-ProCay78} algorithm.

The original paper on this topic by Arthur Cayley \cite{Cayley1846} was published in the \emph{Philosophical Magazine} in 1846. While the work is widely cited and its results are reproduced in numerous modern textbooks and surveys on rotation representations, direct digital access to the original journal issue is limited. For this reason, the citation is retained for historical completeness, with the understanding that the Cayley transform is thoroughly documented in later secondary source (see also modern treatments in \cite{Shuster1993,Hartley2004}).

{\em Properties of skew-symmetric matrices $S$}
\begin{itemize}
\item {\em Defining property}: $S\in\mathbb{R}^{n\times n},\ S=-S^\top$.
\item {\em Zero diagonal}: $\forall i\in[n],\ S_{ii}=0$.
\item {\em Vanishing quadratic form}: $\forall x\in\mathbb{R}^n,\ x^\top S x = 0$.
\item {\em Invertibility of $(I_n-S)$}: $\det(I_n-S)\neq 0$, hence $(I_n-S)$ is invertible.
\item {\em Equality of determinants}: $\det(I_n+S)=\det(I_n-S)$.
\item The determinant of the Cayley matrix $R \eqd (I_n+S)(I_n-S)^{-1}$ equals one.
\item {\em Hat matrix is skew-symmetric}: Let $u\in\mathbb{R}^3$. Then the hat matrix
$$
S_u \eqd \hat{u} \eqd
\begin{bmatrix}
0 & -u_z & u_y\\
u_z & 0 & -u_x\\
-u_y & u_x & 0
\end{bmatrix}
$$
is skew-symmetric. Conversely, if $S\in\mathbb{R}^{3\times3}$ is skew-symmetric, then there exists exactly one vector $u\in\mathbb{R}^3$ such that $\hat{u}=S$.
\item Properties of the hat matrix:
$\hat{u}u=\mathbf{0}_3$, $\hat{u}v=-\hat{v}u$,
$\hat{u}^2=uu^\top-\|u\|^2I_3$.
\end{itemize}

{\em General Cayley formula}
\begin{itemize}
\item {\em Cayley matrix property}: If $S\in\mathbb{R}^{n\times n}$ is skew-symmetric, then
$$
R_S \eqd (I_n+S)(I_n-S)^{-1}
$$
is a rotation matrix. It is widely accepted that Cayley included this general formula in 1846 paper \cite{Cayley1846}.
\item {\em Inverse Cayley theorem}: If $R$ is an orthogonal matrix such that $R+I$ is nonsingular, then
$$
S_R \eqd (R+I)^{-1}(R-I)
$$
is skew-symmetric and satisfies $R_{S_R}=R$.
\end{itemize}

{\em Properties of the Cayley formula in 3D}
\begin{itemize}
\item {\em Auxiliary identity for the 3D Cayley formula}:
$$
v\in\mathbb{R}^3 \Lra
\left[(1-\|v\|^2)I_3 + 2vv^\top + 2\hat{v}\right](I_3-\hat{v})
= (1+\|v\|^2)(I_3+\hat{v}) .
$$
\item {\em Cayley formula for a rotation matrix $R_v$}:
$$
R_v \eqd (I_3+\hat{v})(I_3-\hat{v})^{-1}
\Lra
R_v =
\frac{(1-\|v\|^2)I_3 + 2vv^\top + 2\hat{v}}{1+\|v\|^2}.
$$
In the literature, this formula is also called the Cayley-Rodrigues formula for rotations \cite{Barfoot2017}, although Rodrigues's name refers to the representation of rotations in the Lie group \cite{Ma2004}. However, the vector $v$, which uniquely determines the skew-symmetric matrix $\hat{v}$, is called the Gibbs vector, see Barfoot \cite{Barfoot2017}. 
In this paper, we refer to this vector v as the Cayley vector,
 paying tribute to the creator of this representation for rotations.
\item {\em Equivalent Cayley formulas}:
$$
R_v =
\frac{(1-\|v\|^2)I_3 + 2vv^\top + 2\hat{v}}{1+\|v\|^2}
=
I_3 + \frac{2\hat{v}(I_3+\hat{v})}{1+\|v\|^2}
=
I_3 + \frac{2(I_3+\hat{v})\hat{v}}{1+\|v\|^2}.
$$
\item {\em $v$ lies on the rotation axis}: $R_v v = v$.
\item {\em Negation of the Cayley vector gives the inverse rotation}:
$v\mapsto -v \Lra R_{-v}=R_v^{-1}=R_v^\top$.
\item {\em Zero Cayley vector gives the identity}: $v=\mathbf{0}_3 \Lra R_0=I_3$.
\item {\em Decomposition of the Cayley transform result}:
$$
x\perp v,\quad x^\perp \eqd \frac{v\times x}{\|v\|},\quad
y\eqd R_v x \Lra
y =
\frac{1-\|v\|^2}{1+\|v\|^2}x +
\frac{2\|v\|}{1+\|v\|^2}x^\perp,
\quad
\frac{x^\top y}{\|x\|\|y\|}
\equiv
\frac{1-\|v\|^2}{1+\|v\|^2}.
$$
\item {\em Conclusion}: $R_v$ represents a rotation about axis $v$ by an angle $\varphi$ such that
$$
\cos\varphi=\frac{1-\|v\|^2}{1+\|v\|^2},\quad
\sin\varphi=\pm\frac{2\|v\|}{1+\|v\|^2},\quad
\tan(\varphi/2)=\pm\|v\|.
$$
\end{itemize}

\section{The Rotational Procrustes Problem (Procrustes$^{+}$)}
\label{app:procrustes}

The Procrustes problem is usually formulated as finding the orthonormal matrix $R$ closest (in the Frobenius norm) to a given matrix $A$:
$$
R=\arg\min \|A-R\|_F.
$$
In pose estimation this problem is restricted to the class of rotation matrices, i.e.\ matrices satisfying $\det R=1$.
This rotational Procrustes problem was formulated by Shinji Umeyama~\cite{Umeyama1991}, who showed that in the standard solution
$R=UV^\top$, where $A\xeq{\mathrm{svd}}U\Sigma V^\top$, an optimal correction must be applied to $U$:
$$
A\xeq U\Sigma V^\top,\quad
U=[u_1,u_2,u_3]
\Lra
U'=[u_1,u_2,-u_3],\quad
R'=U'V^\top.
$$
This result is crucial for the \texttt{PnP-ProCay78} algorithm, since the Procrustes$^{+}$ procedure is used to determine the nearest rotation both for the starting points and for their antipodal counterparts.

In this appendix we provide a more elementary proof of the correctness of Umeyama’s correction.

\begin{Theorem}[Rotational Procrustes problem]
\label{th:rot-procrust}
Let $A\in\mathbb{R}^{k\times k}$ and
$$
A \xeq{\mathrm{svd}} U\Sigma V^\top,\qquad
\Sigma=\operatorname{diag}(\sigma_1,\dots,\sigma_k),\quad
\sigma_1>\dots>\sigma_{k-1}\ge 0.
$$
We seek
$$
R^\ast=\arg\min_{R\in\SO{k}}\|A-R\|_F^2.
$$
Then the solution is
$$
R^\ast = U D^\ast V^\top,
$$
where
$$
D^\ast=
\begin{cases}
I_k, & \det(UV^\top)=+1,\\
\operatorname{diag}(1,\dots,1,-1), & \det(UV^\top)=-1.
\end{cases}
$$
If $\operatorname{rank}(A)=k$ or $k-1$, this solution is unique.
\end{Theorem}

\begin{proof}
Let $S=\operatorname{diag}(s_1,\dots,s_k)$ with $s_i\in\{\pm1\}$.
Then $S$ is orthonormal and $S^2=I_k$, hence
$$
A=U\Sigma V^\top=(US)(S\Sigma)V^\top.
$$
There are therefore $2^k$ equivalent SVD decompositions. No more exist because $U$ forms an eigenbasis of $A^\top A$ with eigenvalues $\lambda_i=\sigma_i^2$, each having multiplicity one.

Since multiplication by orthonormal matrices preserves the Frobenius norm,
$$
\|A-R\|_F^2
=\|(US)^\top(A-R)V\|_F^2
=\|S\Sigma-(US)^\top R V\|_F^2.
$$
Define
$$
W\eqd(US)^\top R V\in SO(k),
$$
so that
$$
\|A-R\|_F^2=\|S\Sigma-W\|_F^2.
$$

{\bf Step 1. Reduction to diagonal matrices.}

Expanding the norm,
$$
\|S\Sigma-W\|_F^2
=\sum_{i=1}^k(s_i\sigma_i-w_{ii})^2
+\sum_{i\neq j}w_{ij}^2.
$$
Setting off-diagonal entries of $W$ to zero restricts the search to diagonal orthonormal matrices,
$$
W=\operatorname{diag}(w_{11},\dots,w_{kk}),\quad
w_{ii}\in\{\pm1\}.
$$
Thus every minimizer has this form. The corresponding rotation is recovered as
$$
R=U(SW)V^\top=UDV^\top,\quad D\eqd SW.
$$

{\bf Step 2. Reduction to trace maximization.}

For diagonal $W$,
$$
\|S\Sigma-W\|_F^2
=\sum_{i=1}^k(s_i\sigma_i-w_{ii})^2
=\|A\|_F^2+k-2\sum_{i=1}^k\sigma_i d_{ii}.
$$
Hence minimizing the Frobenius norm is equivalent to maximizing
$\operatorname{tr}(\Sigma D)$.

{\bf Step 3. Determinant constraint.}

Since $R=UDV^\top$,
$$
\det R=\det D\cdot\det(UV^\top)=1,
$$
which implies
$$
\det D=\det(UV^\top).
$$
Thus the signs $d_i\in\{\pm1\}$ must be chosen so that their product has the required sign and $\sum_i\sigma_i d_i$ is maximized.

Because $\sigma_1\ge\dots\ge\sigma_k\ge0$:
\begin{itemize}
\item If $\det(UV^\top)=+1$, the optimum is
$$
D=I_k,\qquad R^\ast=UV^\top.
$$
\item If $\det(UV^\top)=-1$, exactly one negative sign is required, placed at the smallest $\sigma_k$:
$$
D=\operatorname{diag}(1,\dots,1,-1),\qquad
R^\ast=[u_1,\dots,u_{k-1},-u_k]V^\top.
$$
\end{itemize}

{\bf Uniqueness.}
If $\operatorname{rank}(A)=k$ or $k-1$, the smallest singular value is unique, so the position of $-1$ is uniquely determined, yielding a unique optimum. If the rank is lower, multiple optimal solutions exist. This completes the proof.
\end{proof}

\section{Tables of Reconstruction Error Values in the Eigenbasis}

\subsection{Results for RGB Images}

\begin{table}[H]
\centering
\caption{Eigenvalues of the matrix $\Omega^{(rgb)}$ for RGB images. The notation $\lambda_{[i]}^{(\pm k)}$ denotes the value $\lambda_i\cdot 10^{\pm k}$.}
\label{tab:rgb-omega-eigenvalues}
\small
\begin{tabular}{lccccccccc}
\toprule
name &
$\lambda_1$ & $\lambda_2$ & $\lambda_3$ & $\lambda_4$ & $\lambda_5$ &
$\lambda_6$ & $\lambda_7$ & $\lambda_8$ & $\lambda_9$ \\
\midrule
\texttt{2\_rgb} & $0.0_{[1]}^{(+0)}$ & $9.80_{[2]}^{(-35)}$ & $1.76_{[3]}^{(-17)}$ & $4.26_{[4]}^{(-17)}$ & $2.72_{[5]}^{(-4)}$ & $5.49_{[6]}^{(-4)}$ & $1.00_{[7]}^{(+0)}$ & $1.00_{[8]}^{(+0)}$ & $1.00_{[9]}^{(+0)}$ \\
\texttt{3\_rgb} & $0.0_{[1]}^{(+0)}$ & $2.51_{[2]}^{(-34)}$ & $3.66_{[3]}^{(-17)}$ & $1.14_{[4]}^{(-16)}$ & $4.17_{[5]}^{(-4)}$ & $5.27_{[6]}^{(-4)}$ & $1.00_{[7]}^{(+0)}$ & $1.00_{[8]}^{(+0)}$ & $1.00_{[9]}^{(+0)}$ \\
\texttt{4\_rgb} & $0.0_{[1]}^{(+0)}$ & $1.00_{[2]}^{(-36)}$ & $9.57_{[3]}^{(-18)}$ & $1.28_{[4]}^{(-16)}$ & $3.91_{[5]}^{(-4)}$ & $4.13_{[6]}^{(-4)}$ & $1.00_{[7]}^{(+0)}$ & $1.00_{[8]}^{(+0)}$ & $1.00_{[9]}^{(+0)}$ \\
\texttt{5\_rgb} & $0.0_{[1]}^{(+0)}$ & $7.22_{[2]}^{(-35)}$ & $7.01_{[3]}^{(-19)}$ & $9.33_{[4]}^{(-17)}$ & $3.57_{[5]}^{(-4)}$ & $3.59_{[6]}^{(-4)}$ & $1.00_{[7]}^{(+0)}$ & $1.00_{[8]}^{(+0)}$ & $1.00_{[9]}^{(+0)}$ \\
\texttt{6\_rgb} & $0.0_{[1]}^{(+0)}$ & $6.99_{[2]}^{(-35)}$ & $8.47_{[3]}^{(-18)}$ & $4.38_{[4]}^{(-17)}$ & $3.29_{[5]}^{(-4)}$ & $3.62_{[6]}^{(-4)}$ & $1.00_{[7]}^{(+0)}$ & $1.00_{[8]}^{(+0)}$ & $1.00_{[9]}^{(+0)}$ \\
\texttt{7\_rgb} & $0.0_{[1]}^{(+0)}$ & $1.04_{[2]}^{(-34)}$ & $9.60_{[3]}^{(-18)}$ & $3.60_{[4]}^{(-17)}$ & $2.52_{[5]}^{(-4)}$ & $4.41_{[6]}^{(-4)}$ & $1.00_{[7]}^{(+0)}$ & $1.00_{[8]}^{(+0)}$ & $1.00_{[9]}^{(+0)}$ \\
\texttt{8\_rgb} & $0.0_{[1]}^{(+0)}$ & $1.84_{[2]}^{(-34)}$ & $4.63_{[3]}^{(-18)}$ & $1.48_{[4]}^{(-16)}$ & $2.20_{[5]}^{(-4)}$ & $5.43_{[6]}^{(-4)}$ & $1.00_{[7]}^{(+0)}$ & $1.00_{[8]}^{(+0)}$ & $1.00_{[9]}^{(+0)}$ \\
\texttt{9\_rgb} & $0.0_{[1]}^{(+0)}$ & $2.79_{[2]}^{(-36)}$ & $3.12_{[3]}^{(-17)}$ & $1.77_{[4]}^{(-16)}$ & $1.61_{[5]}^{(-4)}$ & $6.66_{[6]}^{(-4)}$ & $1.00_{[7]}^{(+0)}$ & $1.00_{[8]}^{(+0)}$ & $1.00_{[9]}^{(+0)}$ \\
\texttt{10\_rgb} & $0.0_{[1]}^{(+0)}$ & $3.30_{[2]}^{(-35)}$ & $3.52_{[3]}^{(-17)}$ & $5.96_{[4]}^{(-17)}$ & $1.45_{[5]}^{(-4)}$ & $8.23_{[6]}^{(-4)}$ & $1.00_{[7]}^{(+0)}$ & $1.00_{[8]}^{(+0)}$ & $1.00_{[9]}^{(+0)}$ \\
\texttt{14\_rgb} & $0.0_{[1]}^{(+0)}$ & $3.91_{[2]}^{(-34)}$ & $1.15_{[3]}^{(-17)}$ & $5.33_{[4]}^{(-17)}$ & $8.16_{[5]}^{(-5)}$ & $7.22_{[6]}^{(-4)}$ & $1.00_{[7]}^{(+0)}$ & $1.00_{[8]}^{(+0)}$ & $1.00_{[9]}^{(+0)}$ \\
\texttt{15\_rgb} & $0.0_{[1]}^{(+0)}$ & $5.40_{[2]}^{(-34)}$ & $2.58_{[3]}^{(-17)}$ & $8.29_{[4]}^{(-17)}$ & $1.70_{[5]}^{(-4)}$ & $6.85_{[6]}^{(-4)}$ & $1.00_{[7]}^{(+0)}$ & $1.00_{[8]}^{(+0)}$ & $1.00_{[9]}^{(+0)}$ \\
\texttt{16\_rgb} & $0.0_{[1]}^{(+0)}$ & $2.65_{[2]}^{(-34)}$ & $6.41_{[3]}^{(-17)}$ & $7.98_{[4]}^{(-17)}$ & $3.05_{[5]}^{(-4)}$ & $5.51_{[6]}^{(-4)}$ & $1.00_{[7]}^{(+0)}$ & $1.00_{[8]}^{(+0)}$ & $1.00_{[9]}^{(+0)}$ \\
\texttt{17\_rgb} & $0.0_{[1]}^{(+0)}$ & $7.90_{[2]}^{(-34)}$ & $1.38_{[3]}^{(-17)}$ & $1.27_{[4]}^{(-16)}$ & $3.66_{[5]}^{(-4)}$ & $4.79_{[6]}^{(-4)}$ & $1.00_{[7]}^{(+0)}$ & $1.00_{[8]}^{(+0)}$ & $1.00_{[9]}^{(+0)}$ \\
\texttt{18\_rgb} & $0.0_{[1]}^{(+0)}$ & $1.64_{[2]}^{(-33)}$ & $1.91_{[3]}^{(-17)}$ & $6.18_{[4]}^{(-17)}$ & $3.67_{[5]}^{(-4)}$ & $3.86_{[6]}^{(-4)}$ & $1.00_{[7]}^{(+0)}$ & $1.00_{[8]}^{(+0)}$ & $1.00_{[9]}^{(+0)}$ \\
\texttt{19\_rgb} & $0.0_{[1]}^{(+0)}$ & $1.95_{[2]}^{(-33)}$ & $2.01_{[3]}^{(-17)}$ & $4.99_{[4]}^{(-17)}$ & $3.43_{[5]}^{(-4)}$ & $3.69_{[6]}^{(-4)}$ & $1.00_{[7]}^{(+0)}$ & $1.00_{[8]}^{(+0)}$ & $1.00_{[9]}^{(+0)}$ \\
\texttt{20\_rgb} & $0.0_{[1]}^{(+0)}$ & $3.80_{[2]}^{(-34)}$ & $3.79_{[3]}^{(-18)}$ & $6.06_{[4]}^{(-17)}$ & $3.13_{[5]}^{(-4)}$ & $3.34_{[6]}^{(-4)}$ & $1.00_{[7]}^{(+0)}$ & $1.00_{[8]}^{(+0)}$ & $1.00_{[9]}^{(+0)}$ \\
\texttt{21\_rgb} & $0.0_{[1]}^{(+0)}$ & $5.20_{[2]}^{(-34)}$ & $1.94_{[3]}^{(-17)}$ & $5.00_{[4]}^{(-17)}$ & $3.82_{[5]}^{(-4)}$ & $4.02_{[6]}^{(-4)}$ & $1.00_{[7]}^{(+0)}$ & $1.00_{[8]}^{(+0)}$ & $1.00_{[9]}^{(+0)}$ \\
\texttt{22\_rgb} & $0.0_{[1]}^{(+0)}$ & $9.35_{[2]}^{(-34)}$ & $9.90_{[3]}^{(-19)}$ & $7.60_{[4]}^{(-17)}$ & $3.52_{[5]}^{(-4)}$ & $4.71_{[6]}^{(-4)}$ & $1.00_{[7]}^{(+0)}$ & $1.00_{[8]}^{(+0)}$ & $1.00_{[9]}^{(+0)}$ \\
\texttt{23\_rgb} & $0.0_{[1]}^{(+0)}$ & $7.43_{[2]}^{(-36)}$ & $2.84_{[3]}^{(-17)}$ & $1.43_{[4]}^{(-16)}$ & $2.96_{[5]}^{(-4)}$ & $5.75_{[6]}^{(-4)}$ & $1.00_{[7]}^{(+0)}$ & $1.00_{[8]}^{(+0)}$ & $1.00_{[9]}^{(+0)}$ \\
\texttt{32\_rgb} & $0.0_{[1]}^{(+0)}$ & $1.87_{[2]}^{(-34)}$ & $1.48_{[3]}^{(-18)}$ & $5.51_{[4]}^{(-17)}$ & $2.25_{[5]}^{(-4)}$ & $5.34_{[6]}^{(-4)}$ & $1.00_{[7]}^{(+0)}$ & $1.00_{[8]}^{(+0)}$ & $1.00_{[9]}^{(+0)}$ \\
\texttt{34\_rgb} & $0.0_{[1]}^{(+0)}$ & $9.20_{[2]}^{(-34)}$ & $1.89_{[3]}^{(-17)}$ & $1.30_{[4]}^{(-16)}$ & $2.40_{[5]}^{(-4)}$ & $4.31_{[6]}^{(-4)}$ & $1.00_{[7]}^{(+0)}$ & $1.00_{[8]}^{(+0)}$ & $1.00_{[9]}^{(+0)}$ \\
\texttt{35\_rgb} & $0.0_{[1]}^{(+0)}$ & $5.62_{[2]}^{(-34)}$ & $2.22_{[3]}^{(-18)}$ & $6.73_{[4]}^{(-17)}$ & $2.47_{[5]}^{(-4)}$ & $4.00_{[6]}^{(-4)}$ & $1.00_{[7]}^{(+0)}$ & $1.00_{[8]}^{(+0)}$ & $1.00_{[9]}^{(+0)}$ \\
\texttt{41\_rgb} & $0.0_{[1]}^{(+0)}$ & $1.15_{[2]}^{(-33)}$ & $1.88_{[3]}^{(-17)}$ & $6.19_{[4]}^{(-17)}$ & $3.45_{[5]}^{(-4)}$ & $3.91_{[6]}^{(-4)}$ & $1.00_{[7]}^{(+0)}$ & $1.00_{[8]}^{(+0)}$ & $1.00_{[9]}^{(+0)}$ \\
\texttt{42\_rgb} & $0.0_{[1]}^{(+0)}$ & $3.06_{[2]}^{(-34)}$ & $3.07_{[3]}^{(-17)}$ & $7.25_{[4]}^{(-17)}$ & $3.37_{[5]}^{(-4)}$ & $3.88_{[6]}^{(-4)}$ & $1.00_{[7]}^{(+0)}$ & $1.00_{[8]}^{(+0)}$ & $1.00_{[9]}^{(+0)}$ \\
\texttt{43\_rgb} & $0.0_{[1]}^{(+0)}$ & $5.91_{[2]}^{(-34)}$ & $4.66_{[3]}^{(-17)}$ & $8.00_{[4]}^{(-17)}$ & $2.95_{[5]}^{(-4)}$ & $3.59_{[6]}^{(-4)}$ & $1.00_{[7]}^{(+0)}$ & $1.00_{[8]}^{(+0)}$ & $1.00_{[9]}^{(+0)}$ \\
\texttt{44\_rgb} & $0.0_{[1]}^{(+0)}$ & $1.17_{[2]}^{(-34)}$ & $8.34_{[3]}^{(-18)}$ & $1.40_{[4]}^{(-16)}$ & $3.11_{[5]}^{(-4)}$ & $3.61_{[6]}^{(-4)}$ & $1.00_{[7]}^{(+0)}$ & $1.00_{[8]}^{(+0)}$ & $1.00_{[9]}^{(+0)}$ \\
\texttt{45\_rgb} & $0.0_{[1]}^{(+0)}$ & $9.07_{[2]}^{(-34)}$ & $2.72_{[3]}^{(-17)}$ & $1.51_{[4]}^{(-16)}$ & $3.23_{[5]}^{(-4)}$ & $3.66_{[6]}^{(-4)}$ & $1.00_{[7]}^{(+0)}$ & $1.00_{[8]}^{(+0)}$ & $1.00_{[9]}^{(+0)}$ \\
\texttt{46\_rgb} & $0.0_{[1]}^{(+0)}$ & $2.80_{[2]}^{(-35)}$ & $7.95_{[3]}^{(-18)}$ & $1.20_{[4]}^{(-16)}$ & $2.79_{[5]}^{(-4)}$ & $3.85_{[6]}^{(-4)}$ & $1.00_{[7]}^{(+0)}$ & $1.00_{[8]}^{(+0)}$ & $1.00_{[9]}^{(+0)}$ \\
\texttt{53\_rgb} & $0.0_{[1]}^{(+0)}$ & $6.72_{[2]}^{(-35)}$ & $1.35_{[3]}^{(-17)}$ & $2.79_{[4]}^{(-17)}$ & $2.20_{[5]}^{(-4)}$ & $3.47_{[6]}^{(-4)}$ & $1.00_{[7]}^{(+0)}$ & $1.00_{[8]}^{(+0)}$ & $1.00_{[9]}^{(+0)}$ \\
\texttt{54\_rgb} & $0.0_{[1]}^{(+0)}$ & $1.92_{[2]}^{(-34)}$ & $6.69_{[3]}^{(-19)}$ & $9.15_{[4]}^{(-17)}$ & $1.98_{[5]}^{(-4)}$ & $3.51_{[6]}^{(-4)}$ & $1.00_{[7]}^{(+0)}$ & $1.00_{[8]}^{(+0)}$ & $1.00_{[9]}^{(+0)}$ \\
\texttt{56\_rgb} & $0.0_{[1]}^{(+0)}$ & $1.56_{[2]}^{(-33)}$ & $2.18_{[3]}^{(-17)}$ & $6.44_{[4]}^{(-17)}$ & $1.84_{[5]}^{(-4)}$ & $3.69_{[6]}^{(-4)}$ & $1.00_{[7]}^{(+0)}$ & $1.00_{[8]}^{(+0)}$ & $1.00_{[9]}^{(+0)}$ \\
\bottomrule
\end{tabular}
\end{table}

\subsection{Results for IR Images}

\begin{table}[H]
\centering
\caption{Eigenvalues of the matrix $\Omega^{(ir)}$ for thermal images. The notation $\lambda_{[i]}^{(\pm k)}$ denotes the value $\lambda_i\cdot 10^{\pm k}$.}
\label{tab:ir-omega-eigenvalues}
\small
\begin{tabular}{lccccccccc}
\toprule
name &
$\lambda_1$ & $\lambda_2$ & $\lambda_3$ & $\lambda_4$ & $\lambda_5$ &
$\lambda_6$ & $\lambda_7$ & $\lambda_8$ & $\lambda_9$ \\
\midrule
\texttt{2\_ir} & $0.0_{[1]}^{(+0)}$ & $3.08_{[2]}^{(-34)}$ & $2.25_{[3]}^{(-18)}$ & $6.90_{[4]}^{(-17)}$ & $2.01_{[5]}^{(-3)}$ & $3.99_{[6]}^{(-3)}$ & $1.00_{[7]}^{(+0)}$ & $1.00_{[8]}^{(+0)}$ & $1.00_{[9]}^{(+0)}$ \\
\texttt{3\_ir} & $0.0_{[1]}^{(+0)}$ & $1.13_{[2]}^{(-34)}$ & $1.10_{[3]}^{(-17)}$ & $8.70_{[4]}^{(-17)}$ & $2.31_{[5]}^{(-3)}$ & $3.43_{[6]}^{(-3)}$ & $1.00_{[7]}^{(+0)}$ & $1.00_{[8]}^{(+0)}$ & $1.00_{[9]}^{(+0)}$ \\
\texttt{4\_ir} & $0.0_{[1]}^{(+0)}$ & $2.46_{[2]}^{(-33)}$ & $4.17_{[3]}^{(-17)}$ & $8.60_{[4]}^{(-17)}$ & $2.30_{[5]}^{(-3)}$ & $3.69_{[6]}^{(-3)}$ & $1.00_{[7]}^{(+0)}$ & $1.00_{[8]}^{(+0)}$ & $1.00_{[9]}^{(+0)}$ \\
\texttt{5\_ir} & $0.0_{[1]}^{(+0)}$ & $1.17_{[2]}^{(-34)}$ & $3.82_{[3]}^{(-18)}$ & $8.80_{[4]}^{(-17)}$ & $2.11_{[5]}^{(-3)}$ & $2.42_{[6]}^{(-3)}$ & $1.00_{[7]}^{(+0)}$ & $1.00_{[8]}^{(+0)}$ & $1.00_{[9]}^{(+0)}$ \\
\texttt{6\_ir} & $0.0_{[1]}^{(+0)}$ & $1.70_{[2]}^{(-35)}$ & $4.26_{[3]}^{(-18)}$ & $1.65_{[4]}^{(-16)}$ & $2.03_{[5]}^{(-3)}$ & $2.17_{[6]}^{(-3)}$ & $1.00_{[7]}^{(+0)}$ & $1.00_{[8]}^{(+0)}$ & $1.00_{[9]}^{(+0)}$ \\
\texttt{7\_ir} & $0.0_{[1]}^{(+0)}$ & $7.44_{[2]}^{(-36)}$ & $2.46_{[3]}^{(-17)}$ & $3.60_{[4]}^{(-17)}$ & $1.53_{[5]}^{(-3)}$ & $4.60_{[6]}^{(-3)}$ & $1.00_{[7]}^{(+0)}$ & $1.00_{[8]}^{(+0)}$ & $1.00_{[9]}^{(+0)}$ \\
\texttt{16\_ir} & $0.0_{[1]}^{(+0)}$ & $1.53_{[2]}^{(-34)}$ & $2.94_{[3]}^{(-17)}$ & $6.91_{[4]}^{(-17)}$ & $1.85_{[5]}^{(-3)}$ & $3.63_{[6]}^{(-3)}$ & $1.00_{[7]}^{(+0)}$ & $1.00_{[8]}^{(+0)}$ & $1.00_{[9]}^{(+0)}$ \\
\texttt{17\_ir} & $0.0_{[1]}^{(+0)}$ & $1.19_{[2]}^{(-33)}$ & $4.91_{[3]}^{(-19)}$ & $5.04_{[4]}^{(-17)}$ & $2.23_{[5]}^{(-3)}$ & $4.35_{[6]}^{(-3)}$ & $1.00_{[7]}^{(+0)}$ & $1.00_{[8]}^{(+0)}$ & $1.00_{[9]}^{(+0)}$ \\
\texttt{18\_ir} & $0.0_{[1]}^{(+0)}$ & $1.26_{[2]}^{(-34)}$ & $5.19_{[3]}^{(-18)}$ & $1.01_{[4]}^{(-16)}$ & $2.37_{[5]}^{(-3)}$ & $2.76_{[6]}^{(-3)}$ & $1.00_{[7]}^{(+0)}$ & $1.00_{[8]}^{(+0)}$ & $1.00_{[9]}^{(+0)}$ \\
\texttt{19\_ir} & $0.0_{[1]}^{(+0)}$ & $4.69_{[2]}^{(-34)}$ & $4.32_{[3]}^{(-17)}$ & $1.04_{[4]}^{(-16)}$ & $1.86_{[5]}^{(-3)}$ & $3.15_{[6]}^{(-3)}$ & $1.00_{[7]}^{(+0)}$ & $1.00_{[8]}^{(+0)}$ & $1.00_{[9]}^{(+0)}$ \\
\texttt{20\_ir} & $0.0_{[1]}^{(+0)}$ & $1.80_{[2]}^{(-33)}$ & $1.82_{[3]}^{(-17)}$ & $3.70_{[4]}^{(-17)}$ & $2.14_{[5]}^{(-3)}$ & $2.21_{[6]}^{(-3)}$ & $1.00_{[7]}^{(+0)}$ & $1.00_{[8]}^{(+0)}$ & $1.00_{[9]}^{(+0)}$ \\
\texttt{21\_ir} & $0.0_{[1]}^{(+0)}$ & $1.98_{[2]}^{(-34)}$ & $2.07_{[3]}^{(-19)}$ & $6.28_{[4]}^{(-17)}$ & $1.81_{[5]}^{(-3)}$ & $3.66_{[6]}^{(-3)}$ & $1.00_{[7]}^{(+0)}$ & $1.00_{[8]}^{(+0)}$ & $1.00_{[9]}^{(+0)}$ \\
\texttt{22\_ir} & $0.0_{[1]}^{(+0)}$ & $6.09_{[2]}^{(-34)}$ & $8.77_{[3]}^{(-18)}$ & $1.03_{[4]}^{(-16)}$ & $2.03_{[5]}^{(-3)}$ & $2.76_{[6]}^{(-3)}$ & $1.00_{[7]}^{(+0)}$ & $1.00_{[8]}^{(+0)}$ & $1.00_{[9]}^{(+0)}$ \\
\texttt{23\_ir} & $0.0_{[1]}^{(+0)}$ & $1.00_{[2]}^{(-33)}$ & $2.32_{[3]}^{(-17)}$ & $1.11_{[4]}^{(-16)}$ & $2.85_{[5]}^{(-3)}$ & $3.18_{[6]}^{(-3)}$ & $1.00_{[7]}^{(+0)}$ & $1.00_{[8]}^{(+0)}$ & $1.00_{[9]}^{(+0)}$ \\
\texttt{31\_ir} & $0.0_{[1]}^{(+0)}$ & $4.31_{[2]}^{(-33)}$ & $2.98_{[3]}^{(-17)}$ & $8.03_{[4]}^{(-17)}$ & $1.13_{[5]}^{(-3)}$ & $2.61_{[6]}^{(-3)}$ & $1.00_{[7]}^{(+0)}$ & $1.00_{[8]}^{(+0)}$ & $1.00_{[9]}^{(+0)}$ \\
\texttt{32\_ir} & $0.0_{[1]}^{(+0)}$ & $3.81_{[2]}^{(-34)}$ & $9.65_{[3]}^{(-18)}$ & $9.99_{[4]}^{(-17)}$ & $1.69_{[5]}^{(-3)}$ & $1.81_{[6]}^{(-3)}$ & $1.00_{[7]}^{(+0)}$ & $1.00_{[8]}^{(+0)}$ & $1.00_{[9]}^{(+0)}$ \\
\texttt{34\_ir} & $0.0_{[1]}^{(+0)}$ & $1.04_{[2]}^{(-33)}$ & $3.45_{[3]}^{(-17)}$ & $2.32_{[4]}^{(-16)}$ & $1.75_{[5]}^{(-3)}$ & $2.43_{[6]}^{(-3)}$ & $1.00_{[7]}^{(+0)}$ & $1.00_{[8]}^{(+0)}$ & $1.00_{[9]}^{(+0)}$ \\
\texttt{35\_ir} & $0.0_{[1]}^{(+0)}$ & $2.57_{[2]}^{(-33)}$ & $5.43_{[3]}^{(-18)}$ & $3.22_{[4]}^{(-17)}$ & $1.16_{[5]}^{(-3)}$ & $2.10_{[6]}^{(-3)}$ & $1.00_{[7]}^{(+0)}$ & $1.00_{[8]}^{(+0)}$ & $1.00_{[9]}^{(+0)}$ \\
\texttt{41\_ir} & $0.0_{[1]}^{(+0)}$ & $1.81_{[2]}^{(-33)}$ & $1.70_{[3]}^{(-17)}$ & $9.85_{[4]}^{(-17)}$ & $1.80_{[5]}^{(-3)}$ & $3.19_{[6]}^{(-3)}$ & $1.00_{[7]}^{(+0)}$ & $1.00_{[8]}^{(+0)}$ & $1.00_{[9]}^{(+0)}$ \\
\texttt{43\_ir} & $0.0_{[1]}^{(+0)}$ & $4.61_{[2]}^{(-34)}$ & $1.85_{[3]}^{(-18)}$ & $2.05_{[4]}^{(-16)}$ & $2.24_{[5]}^{(-3)}$ & $2.69_{[6]}^{(-3)}$ & $1.00_{[7]}^{(+0)}$ & $1.00_{[8]}^{(+0)}$ & $1.00_{[9]}^{(+0)}$ \\
\texttt{44\_ir} & $0.0_{[1]}^{(+0)}$ & $1.23_{[2]}^{(-34)}$ & $4.05_{[3]}^{(-18)}$ & $5.35_{[4]}^{(-17)}$ & $1.82_{[5]}^{(-3)}$ & $1.97_{[6]}^{(-3)}$ & $1.00_{[7]}^{(+0)}$ & $1.00_{[8]}^{(+0)}$ & $1.00_{[9]}^{(+0)}$ \\
\texttt{45\_ir} & $0.0_{[1]}^{(+0)}$ & $3.71_{[2]}^{(-33)}$ & $2.61_{[3]}^{(-17)}$ & $7.73_{[4]}^{(-17)}$ & $1.74_{[5]}^{(-3)}$ & $2.41_{[6]}^{(-3)}$ & $1.00_{[7]}^{(+0)}$ & $1.00_{[8]}^{(+0)}$ & $1.00_{[9]}^{(+0)}$ \\
\bottomrule
\end{tabular}
\end{table}

\section{Tables of Reconstruction Error Values in the Canonical Basis}
\label{app:canonical-error-values}

\subsection{Results for RGB Images}

\begin{table}[H]
\centering
\caption{Diagonal values $\omega\eqd\diag{\Omega^{(rgb)}}$ of the matrix $\Omega^{(rgb)}$ for RGB images. The notation $value_{[i]}^{(\pm k)}$ denotes $value\cdot 10^{\pm k}$, which is equal to $\tp{e}_{\pi_i}\Omega^{(rgb)}e_{\pi_i}$, where $\pi$ is the permutation corresponding to ascending sorting.}
\label{tab:rgb-omega-diagonal}
\small
\begin{tabular}{lccccccccc}
\toprule
name &
$\omega_{\pi_1}$ & $\omega_{\pi_2}$ & $\omega_{\pi_3}$ & $\omega_{\pi_4}$ & $\omega_{\pi_5}$ &
$\omega_{\pi_6}$ & $\omega_{\pi_7}$ & $\omega_{\pi_8}$ & $\omega_{\pi_9}$ \\
\midrule
\texttt{2\_rgb} & $0.0_{[3]}^{(+0)}$ & $0.0_{[6]}^{(+0)}$ & $0.0_{[9]}^{(+0)}$ & $4.91_{[8]}^{(-4)}$ & $6.59_{[7]}^{(-4)}$ & $3.92_{[5]}^{(-1)}$ & $6.12_{[1]}^{(-1)}$ & $9.96_{[2]}^{(-1)}$ & $1.00_{[4]}^{(+0)}$ \\
\texttt{3\_rgb} & $0.0_{[3]}^{(+0)}$ & $0.0_{[6]}^{(+0)}$ & $0.0_{[9]}^{(+0)}$ & $4.30_{[8]}^{(-4)}$ & $5.26_{[7]}^{(-4)}$ & $4.62_{[5]}^{(-1)}$ & $5.42_{[1]}^{(-1)}$ & $9.98_{[2]}^{(-1)}$ & $9.99_{[4]}^{(-1)}$ \\
\texttt{4\_rgb} & $0.0_{[3]}^{(+0)}$ & $0.0_{[6]}^{(+0)}$ & $0.0_{[9]}^{(+0)}$ & $6.71_{[7]}^{(-4)}$ & $7.97_{[8]}^{(-4)}$ & $4.98_{[5]}^{(-1)}$ & $5.05_{[1]}^{(-1)}$ & $9.98_{[2]}^{(-1)}$ & $9.99_{[4]}^{(-1)}$ \\
\texttt{5\_rgb} & $0.0_{[3]}^{(+0)}$ & $0.0_{[6]}^{(+0)}$ & $0.0_{[9]}^{(+0)}$ & $4.16_{[8]}^{(-4)}$ & $4.77_{[7]}^{(-4)}$ & $5.01_{[5]}^{(-1)}$ & $5.02_{[1]}^{(-1)}$ & $9.98_{[4]}^{(-1)}$ & $9.98_{[2]}^{(-1)}$ \\
\texttt{6\_rgb} & $0.0_{[3]}^{(+0)}$ & $0.0_{[6]}^{(+0)}$ & $0.0_{[9]}^{(+0)}$ & $4.73_{[7]}^{(-4)}$ & $4.84_{[8]}^{(-4)}$ & $4.82_{[5]}^{(-1)}$ & $5.22_{[1]}^{(-1)}$ & $9.98_{[4]}^{(-1)}$ & $9.99_{[2]}^{(-1)}$ \\
\texttt{7\_rgb} & $0.0_{[3]}^{(+0)}$ & $0.0_{[6]}^{(+0)}$ & $0.0_{[9]}^{(+0)}$ & $3.87_{[8]}^{(-4)}$ & $4.89_{[7]}^{(-4)}$ & $4.01_{[5]}^{(-1)}$ & $6.03_{[1]}^{(-1)}$ & $9.97_{[4]}^{(-1)}$ & $9.99_{[2]}^{(-1)}$ \\
\texttt{8\_rgb} & $0.0_{[3]}^{(+0)}$ & $0.0_{[6]}^{(+0)}$ & $0.0_{[9]}^{(+0)}$ & $2.39_{[8]}^{(-4)}$ & $4.44_{[7]}^{(-4)}$ & $3.59_{[5]}^{(-1)}$ & $6.45_{[1]}^{(-1)}$ & $9.97_{[4]}^{(-1)}$ & $9.99_{[2]}^{(-1)}$ \\
\texttt{9\_rgb} & $0.0_{[3]}^{(+0)}$ & $0.0_{[6]}^{(+0)}$ & $0.0_{[9]}^{(+0)}$ & $1.80_{[8]}^{(-4)}$ & $4.60_{[7]}^{(-4)}$ & $2.76_{[5]}^{(-1)}$ & $7.28_{[1]}^{(-1)}$ & $9.97_{[4]}^{(-1)}$ & $9.99_{[2]}^{(-1)}$ \\
\texttt{10\_rgb} & $0.0_{[3]}^{(+0)}$ & $0.0_{[6]}^{(+0)}$ & $0.0_{[9]}^{(+0)}$ & $1.72_{[8]}^{(-4)}$ & $5.23_{[7]}^{(-4)}$ & $2.23_{[5]}^{(-1)}$ & $7.81_{[1]}^{(-1)}$ & $9.97_{[4]}^{(-1)}$ & $9.99_{[2]}^{(-1)}$ \\
\texttt{14\_rgb} & $0.0_{[3]}^{(+0)}$ & $0.0_{[6]}^{(+0)}$ & $0.0_{[9]}^{(+0)}$ & $1.90_{[8]}^{(-4)}$ & $7.19_{[7]}^{(-4)}$ & $2.14_{[5]}^{(-1)}$ & $8.33_{[1]}^{(-1)}$ & $9.54_{[4]}^{(-1)}$ & $1.00_{[2]}^{(+0)}$ \\
\texttt{15\_rgb} & $0.0_{[3]}^{(+0)}$ & $0.0_{[6]}^{(+0)}$ & $0.0_{[9]}^{(+0)}$ & $2.20_{[8]}^{(-4)}$ & $5.13_{[7]}^{(-4)}$ & $3.07_{[5]}^{(-1)}$ & $7.24_{[1]}^{(-1)}$ & $9.70_{[4]}^{(-1)}$ & $1.00_{[2]}^{(+0)}$ \\
\texttt{16\_rgb} & $0.0_{[3]}^{(+0)}$ & $0.0_{[6]}^{(+0)}$ & $0.0_{[9]}^{(+0)}$ & $4.75_{[8]}^{(-4)}$ & $8.89_{[7]}^{(-4)}$ & $3.94_{[5]}^{(-1)}$ & $6.18_{[1]}^{(-1)}$ & $9.88_{[4]}^{(-1)}$ & $1.00_{[2]}^{(+0)}$ \\
\texttt{17\_rgb} & $0.0_{[3]}^{(+0)}$ & $0.0_{[6]}^{(+0)}$ & $0.0_{[9]}^{(+0)}$ & $4.72_{[8]}^{(-4)}$ & $6.76_{[7]}^{(-4)}$ & $4.52_{[5]}^{(-1)}$ & $5.59_{[1]}^{(-1)}$ & $9.90_{[4]}^{(-1)}$ & $1.00_{[2]}^{(+0)}$ \\
\texttt{18\_rgb} & $0.0_{[3]}^{(+0)}$ & $0.0_{[6]}^{(+0)}$ & $0.0_{[9]}^{(+0)}$ & $3.54_{[8]}^{(-4)}$ & $3.64_{[7]}^{(-4)}$ & $4.87_{[5]}^{(-1)}$ & $5.18_{[1]}^{(-1)}$ & $9.96_{[4]}^{(-1)}$ & $1.00_{[2]}^{(+0)}$ \\
\texttt{19\_rgb} & $0.0_{[3]}^{(+0)}$ & $0.0_{[6]}^{(+0)}$ & $0.0_{[9]}^{(+0)}$ & $3.61_{[7]}^{(-4)}$ & $3.89_{[8]}^{(-4)}$ & $4.95_{[1]}^{(-1)}$ & $5.06_{[5]}^{(-1)}$ & $9.99_{[4]}^{(-1)}$ & $1.00_{[2]}^{(+0)}$ \\
\texttt{20\_rgb} & $0.0_{[3]}^{(+0)}$ & $0.0_{[6]}^{(+0)}$ & $0.0_{[9]}^{(+0)}$ & $4.73_{[7]}^{(-4)}$ & $4.75_{[8]}^{(-4)}$ & $4.84_{[1]}^{(-1)}$ & $5.16_{[5]}^{(-1)}$ & $1.00_{[4]}^{(+0)}$ & $1.00_{[2]}^{(+0)}$ \\
\texttt{21\_rgb} & $0.0_{[3]}^{(+0)}$ & $0.0_{[6]}^{(+0)}$ & $0.0_{[9]}^{(+0)}$ & $5.02_{[7]}^{(-4)}$ & $5.46_{[8]}^{(-4)}$ & $5.00_{[5]}^{(-1)}$ & $5.02_{[1]}^{(-1)}$ & $9.97_{[4]}^{(-1)}$ & $1.00_{[2]}^{(+0)}$ \\
\texttt{22\_rgb} & $0.0_{[3]}^{(+0)}$ & $0.0_{[6]}^{(+0)}$ & $0.0_{[9]}^{(+0)}$ & $5.38_{[8]}^{(-4)}$ & $6.80_{[7]}^{(-4)}$ & $4.60_{[5]}^{(-1)}$ & $5.48_{[1]}^{(-1)}$ & $9.92_{[4]}^{(-1)}$ & $1.00_{[2]}^{(+0)}$ \\
\texttt{23\_rgb} & $0.0_{[3]}^{(+0)}$ & $0.0_{[6]}^{(+0)}$ & $0.0_{[9]}^{(+0)}$ & $4.36_{[8]}^{(-4)}$ & $8.00_{[7]}^{(-4)}$ & $3.92_{[5]}^{(-1)}$ & $6.20_{[1]}^{(-1)}$ & $9.88_{[4]}^{(-1)}$ & $1.00_{[2]}^{(+0)}$ \\
\texttt{32\_rgb} & $0.0_{[3]}^{(+0)}$ & $0.0_{[6]}^{(+0)}$ & $0.0_{[9]}^{(+0)}$ & $3.98_{[7]}^{(-4)}$ & $6.39_{[8]}^{(-4)}$ & $3.73_{[1]}^{(-1)}$ & $6.29_{[5]}^{(-1)}$ & $9.99_{[2]}^{(-1)}$ & $9.99_{[4]}^{(-1)}$ \\
\texttt{34\_rgb} & $0.0_{[3]}^{(+0)}$ & $0.0_{[6]}^{(+0)}$ & $0.0_{[9]}^{(+0)}$ & $2.89_{[7]}^{(-4)}$ & $4.64_{[8]}^{(-4)}$ & $4.08_{[1]}^{(-1)}$ & $5.94_{[5]}^{(-1)}$ & $9.98_{[4]}^{(-1)}$ & $1.00_{[2]}^{(+0)}$ \\
\texttt{35\_rgb} & $0.0_{[3]}^{(+0)}$ & $0.0_{[6]}^{(+0)}$ & $0.0_{[9]}^{(+0)}$ & $5.03_{[7]}^{(-4)}$ & $5.86_{[8]}^{(-4)}$ & $4.34_{[1]}^{(-1)}$ & $5.82_{[5]}^{(-1)}$ & $9.85_{[4]}^{(-1)}$ & $9.99_{[2]}^{(-1)}$ \\
\texttt{41\_rgb} & $0.0_{[3]}^{(+0)}$ & $0.0_{[6]}^{(+0)}$ & $0.0_{[9]}^{(+0)}$ & $3.89_{[8]}^{(-4)}$ & $4.54_{[7]}^{(-4)}$ & $4.92_{[5]}^{(-1)}$ & $5.14_{[1]}^{(-1)}$ & $9.96_{[4]}^{(-1)}$ & $9.98_{[2]}^{(-1)}$ \\
\texttt{42\_rgb} & $0.0_{[3]}^{(+0)}$ & $0.0_{[6]}^{(+0)}$ & $0.0_{[9]}^{(+0)}$ & $5.84_{[8]}^{(-4)}$ & $6.76_{[7]}^{(-4)}$ & $4.81_{[1]}^{(-1)}$ & $5.21_{[5]}^{(-1)}$ & $9.98_{[2]}^{(-1)}$ & $1.00_{[4]}^{(+0)}$ \\
\texttt{43\_rgb} & $0.0_{[3]}^{(+0)}$ & $0.0_{[6]}^{(+0)}$ & $0.0_{[9]}^{(+0)}$ & $3.41_{[7]}^{(-4)}$ & $3.77_{[8]}^{(-4)}$ & $4.71_{[1]}^{(-1)}$ & $5.33_{[5]}^{(-1)}$ & $9.98_{[2]}^{(-1)}$ & $9.98_{[4]}^{(-1)}$ \\
\texttt{44\_rgb} & $0.0_{[3]}^{(+0)}$ & $0.0_{[6]}^{(+0)}$ & $0.0_{[9]}^{(+0)}$ & $5.41_{[7]}^{(-4)}$ & $7.02_{[8]}^{(-4)}$ & $4.78_{[1]}^{(-1)}$ & $5.33_{[5]}^{(-1)}$ & $9.91_{[4]}^{(-1)}$ & $9.97_{[2]}^{(-1)}$ \\
\texttt{45\_rgb} & $0.0_{[3]}^{(+0)}$ & $0.0_{[6]}^{(+0)}$ & $0.0_{[9]}^{(+0)}$ & $5.69_{[8]}^{(-4)}$ & $7.03_{[7]}^{(-4)}$ & $4.96_{[5]}^{(-1)}$ & $5.29_{[1]}^{(-1)}$ & $9.78_{[4]}^{(-1)}$ & $9.97_{[2]}^{(-1)}$ \\
\texttt{46\_rgb} & $0.0_{[3]}^{(+0)}$ & $0.0_{[6]}^{(+0)}$ & $0.0_{[9]}^{(+0)}$ & $5.40_{[8]}^{(-4)}$ & $6.59_{[7]}^{(-4)}$ & $4.61_{[5]}^{(-1)}$ & $5.79_{[1]}^{(-1)}$ & $9.63_{[4]}^{(-1)}$ & $9.96_{[2]}^{(-1)}$ \\
\texttt{53\_rgb} & $0.0_{[3]}^{(+0)}$ & $0.0_{[6]}^{(+0)}$ & $0.0_{[9]}^{(+0)}$ & $3.65_{[7]}^{(-4)}$ & $4.20_{[8]}^{(-4)}$ & $4.61_{[1]}^{(-1)}$ & $5.89_{[5]}^{(-1)}$ & $9.52_{[4]}^{(-1)}$ & $9.98_{[2]}^{(-1)}$ \\
\texttt{54\_rgb} & $0.0_{[3]}^{(+0)}$ & $0.0_{[6]}^{(+0)}$ & $0.0_{[9]}^{(+0)}$ & $4.41_{[7]}^{(-4)}$ & $5.40_{[8]}^{(-4)}$ & $4.09_{[1]}^{(-1)}$ & $5.98_{[5]}^{(-1)}$ & $9.95_{[4]}^{(-1)}$ & $9.99_{[2]}^{(-1)}$ \\
\texttt{56\_rgb} & $0.0_{[3]}^{(+0)}$ & $0.0_{[6]}^{(+0)}$ & $0.0_{[9]}^{(+0)}$ & $3.93_{[7]}^{(-4)}$ & $5.32_{[8]}^{(-4)}$ & $4.95_{[1]}^{(-1)}$ & $5.61_{[5]}^{(-1)}$ & $9.45_{[4]}^{(-1)}$ & $9.99_{[2]}^{(-1)}$ \\
\bottomrule
\end{tabular}
\end{table}

\subsection{Results for IR Images}

\begin{table}[H]
\centering
\caption{Diagonal values $\omega\eqd\diag{\Omega^{(ir)}}$ of the matrix $\Omega^{(ir)}$ for thermal images. The notation $value_{[i]}^{(\pm k)}$ denotes $value\cdot 10^{\pm k}$, which is equal to $\tp{e}_{\pi_i}\Omega^{(ir)}e_{\pi_i}$, where $\pi$ is the permutation corresponding to ascending sorting.}
\label{tab:ir-omega-diagonal}
\small
\begin{tabular}{lccccccccc}
\toprule
name &
$\omega_{\pi_1}$ & $\omega_{\pi_2}$ & $\omega_{\pi_3}$ & $\omega_{\pi_4}$ & $\omega_{\pi_5}$ &
$\omega_{\pi_6}$ & $\omega_{\pi_7}$ & $\omega_{\pi_8}$ & $\omega_{\pi_9}$ \\
\midrule
\texttt{2\_ir} & $0.0_{[3]}^{(+0)}$ & $0.0_{[6]}^{(+0)}$ & $0.0_{[9]}^{(+0)}$ & $2.37_{[8]}^{(-3)}$ & $5.45_{[7]}^{(-3)}$ & $3.40_{[5]}^{(-1)}$ & $6.84_{[1]}^{(-1)}$ & $9.80_{[2]}^{(-1)}$ & $9.98_{[4]}^{(-1)}$ \\
\texttt{3\_ir} & $0.0_{[3]}^{(+0)}$ & $0.0_{[6]}^{(+0)}$ & $0.0_{[9]}^{(+0)}$ & $4.73_{[8]}^{(-3)}$ & $6.12_{[7]}^{(-3)}$ & $4.39_{[5]}^{(-1)}$ & $5.61_{[1]}^{(-1)}$ & $1.00_{[4]}^{(+0)}$ & $1.00_{[2]}^{(+0)}$ \\
\texttt{4\_ir} & $0.0_{[3]}^{(+0)}$ & $0.0_{[6]}^{(+0)}$ & $0.0_{[9]}^{(+0)}$ & $4.24_{[8]}^{(-3)}$ & $5.10_{[7]}^{(-3)}$ & $4.98_{[5]}^{(-1)}$ & $5.04_{[1]}^{(-1)}$ & $1.00_{[2]}^{(+0)}$ & $1.00_{[4]}^{(+0)}$ \\
\texttt{5\_ir} & $0.0_{[3]}^{(+0)}$ & $0.0_{[6]}^{(+0)}$ & $0.0_{[9]}^{(+0)}$ & $4.58_{[8]}^{(-3)}$ & $4.89_{[7]}^{(-3)}$ & $4.98_{[5]}^{(-1)}$ & $5.03_{[1]}^{(-1)}$ & $1.00_{[4]}^{(+0)}$ & $1.00_{[2]}^{(+0)}$ \\
\texttt{6\_ir} & $0.0_{[3]}^{(+0)}$ & $0.0_{[6]}^{(+0)}$ & $0.0_{[9]}^{(+0)}$ & $3.29_{[8]}^{(-3)}$ & $4.12_{[7]}^{(-3)}$ & $4.90_{[1]}^{(-1)}$ & $5.12_{[5]}^{(-1)}$ & $9.99_{[2]}^{(-1)}$ & $9.99_{[4]}^{(-1)}$ \\
\texttt{7\_ir} & $0.0_{[3]}^{(+0)}$ & $0.0_{[6]}^{(+0)}$ & $0.0_{[9]}^{(+0)}$ & $2.61_{[8]}^{(-3)}$ & $4.49_{[7]}^{(-3)}$ & $3.41_{[5]}^{(-1)}$ & $6.63_{[1]}^{(-1)}$ & $9.99_{[4]}^{(-1)}$ & $1.00_{[2]}^{(+0)}$ \\
\texttt{16\_ir} & $0.0_{[3]}^{(+0)}$ & $0.0_{[6]}^{(+0)}$ & $0.0_{[9]}^{(+0)}$ & $2.87_{[8]}^{(-3)}$ & $4.73_{[7]}^{(-3)}$ & $4.35_{[5]}^{(-1)}$ & $5.87_{[1]}^{(-1)}$ & $9.82_{[4]}^{(-1)}$ & $9.97_{[2]}^{(-1)}$ \\
\texttt{17\_ir} & $0.0_{[3]}^{(+0)}$ & $0.0_{[6]}^{(+0)}$ & $0.0_{[9]}^{(+0)}$ & $3.10_{[8]}^{(-3)}$ & $4.64_{[7]}^{(-3)}$ & $4.37_{[5]}^{(-1)}$ & $5.70_{[1]}^{(-1)}$ & $9.95_{[4]}^{(-1)}$ & $1.00_{[2]}^{(+0)}$ \\
\texttt{18\_ir} & $0.0_{[3]}^{(+0)}$ & $0.0_{[6]}^{(+0)}$ & $0.0_{[9]}^{(+0)}$ & $2.48_{[8]}^{(-3)}$ & $2.54_{[7]}^{(-3)}$ & $4.97_{[1]}^{(-1)}$ & $5.08_{[5]}^{(-1)}$ & $9.98_{[4]}^{(-1)}$ & $9.98_{[2]}^{(-1)}$ \\
\texttt{19\_ir} & $0.0_{[3]}^{(+0)}$ & $0.0_{[6]}^{(+0)}$ & $0.0_{[9]}^{(+0)}$ & $2.50_{[7]}^{(-3)}$ & $2.92_{[8]}^{(-3)}$ & $4.62_{[1]}^{(-1)}$ & $5.40_{[5]}^{(-1)}$ & $9.99_{[2]}^{(-1)}$ & $1.00_{[4]}^{(+0)}$ \\
\texttt{20\_ir} & $0.0_{[3]}^{(+0)}$ & $0.0_{[6]}^{(+0)}$ & $0.0_{[9]}^{(+0)}$ & $2.31_{[7]}^{(-3)}$ & $2.64_{[8]}^{(-3)}$ & $4.90_{[1]}^{(-1)}$ & $5.12_{[5]}^{(-1)}$ & $9.98_{[4]}^{(-1)}$ & $1.00_{[2]}^{(+0)}$ \\
\texttt{21\_ir} & $0.0_{[3]}^{(+0)}$ & $0.0_{[6]}^{(+0)}$ & $0.0_{[9]}^{(+0)}$ & $2.15_{[8]}^{(-3)}$ & $3.04_{[7]}^{(-3)}$ & $4.40_{[5]}^{(-1)}$ & $5.72_{[1]}^{(-1)}$ & $9.90_{[4]}^{(-1)}$ & $9.99_{[2]}^{(-1)}$ \\
\texttt{22\_ir} & $0.0_{[3]}^{(+0)}$ & $0.0_{[6]}^{(+0)}$ & $0.0_{[9]}^{(+0)}$ & $2.27_{[8]}^{(-3)}$ & $2.81_{[7]}^{(-3)}$ & $4.66_{[5]}^{(-1)}$ & $5.53_{[1]}^{(-1)}$ & $9.81_{[4]}^{(-1)}$ & $1.00_{[2]}^{(+0)}$ \\
\texttt{23\_ir} & $0.0_{[3]}^{(+0)}$ & $0.0_{[6]}^{(+0)}$ & $0.0_{[9]}^{(+0)}$ & $2.25_{[8]}^{(-3)}$ & $3.99_{[7]}^{(-3)}$ & $3.46_{[5]}^{(-1)}$ & $6.70_{[1]}^{(-1)}$ & $9.86_{[4]}^{(-1)}$ & $1.00_{[2]}^{(+0)}$ \\
\texttt{31\_ir} & $0.0_{[3]}^{(+0)}$ & $0.0_{[6]}^{(+0)}$ & $0.0_{[9]}^{(+0)}$ & $1.49_{[7]}^{(-3)}$ & $3.44_{[8]}^{(-3)}$ & $3.95_{[1]}^{(-1)}$ & $6.53_{[5]}^{(-1)}$ & $9.53_{[4]}^{(-1)}$ & $1.00_{[2]}^{(+0)}$ \\
\texttt{32\_ir} & $0.0_{[3]}^{(+0)}$ & $0.0_{[6]}^{(+0)}$ & $0.0_{[9]}^{(+0)}$ & $1.77_{[7]}^{(-3)}$ & $3.15_{[8]}^{(-3)}$ & $3.52_{[1]}^{(-1)}$ & $6.50_{[5]}^{(-1)}$ & $9.99_{[4]}^{(-1)}$ & $1.00_{[2]}^{(+0)}$ \\
\texttt{34\_ir} & $0.0_{[3]}^{(+0)}$ & $0.0_{[6]}^{(+0)}$ & $0.0_{[9]}^{(+0)}$ & $1.96_{[7]}^{(-3)}$ & $3.03_{[8]}^{(-3)}$ & $4.18_{[1]}^{(-1)}$ & $5.83_{[5]}^{(-1)}$ & $1.00_{[2]}^{(+0)}$ & $1.00_{[4]}^{(+0)}$ \\
\texttt{35\_ir} & $0.0_{[3]}^{(+0)}$ & $0.0_{[6]}^{(+0)}$ & $0.0_{[9]}^{(+0)}$ & $1.25_{[7]}^{(-3)}$ & $2.07_{[8]}^{(-3)}$ & $3.79_{[1]}^{(-1)}$ & $6.27_{[5]}^{(-1)}$ & $9.94_{[4]}^{(-1)}$ & $1.00_{[2]}^{(+0)}$ \\
\texttt{41\_ir} & $0.0_{[3]}^{(+0)}$ & $0.0_{[6]}^{(+0)}$ & $0.0_{[9]}^{(+0)}$ & $2.30_{[8]}^{(-3)}$ & $2.61_{[7]}^{(-3)}$ & $4.76_{[5]}^{(-1)}$ & $5.32_{[1]}^{(-1)}$ & $9.93_{[4]}^{(-1)}$ & $1.00_{[2]}^{(+0)}$ \\
\texttt{43\_ir} & $0.0_{[3]}^{(+0)}$ & $0.0_{[6]}^{(+0)}$ & $0.0_{[9]}^{(+0)}$ & $3.95_{[7]}^{(-3)}$ & $5.13_{[8]}^{(-3)}$ & $4.05_{[1]}^{(-1)}$ & $5.96_{[5]}^{(-1)}$ & $1.00_{[4]}^{(+0)}$ & $1.00_{[2]}^{(+0)}$ \\
\texttt{44\_ir} & $0.0_{[3]}^{(+0)}$ & $0.0_{[6]}^{(+0)}$ & $0.0_{[9]}^{(+0)}$ & $3.32_{[8]}^{(-3)}$ & $4.08_{[7]}^{(-3)}$ & $4.85_{[5]}^{(-1)}$ & $5.26_{[1]}^{(-1)}$ & $9.94_{[4]}^{(-1)}$ & $9.95_{[2]}^{(-1)}$ \\
\texttt{45\_ir} & $0.0_{[3]}^{(+0)}$ & $0.0_{[6]}^{(+0)}$ & $0.0_{[9]}^{(+0)}$ & $2.61_{[8]}^{(-3)}$ & $3.67_{[7]}^{(-3)}$ & $4.85_{[5]}^{(-1)}$ & $5.25_{[1]}^{(-1)}$ & $9.91_{[4]}^{(-1)}$ & $1.00_{[2]}^{(+0)}$ \\
\bottomrule
\end{tabular}
\end{table}

\section{State Trajectories in Cayley Space -- Additional Experiments}
\label{app:traj-05-20}

\begin{figure}[H]
\begin{tabular}{cc}
\includegraphics[width=0.48\linewidth]{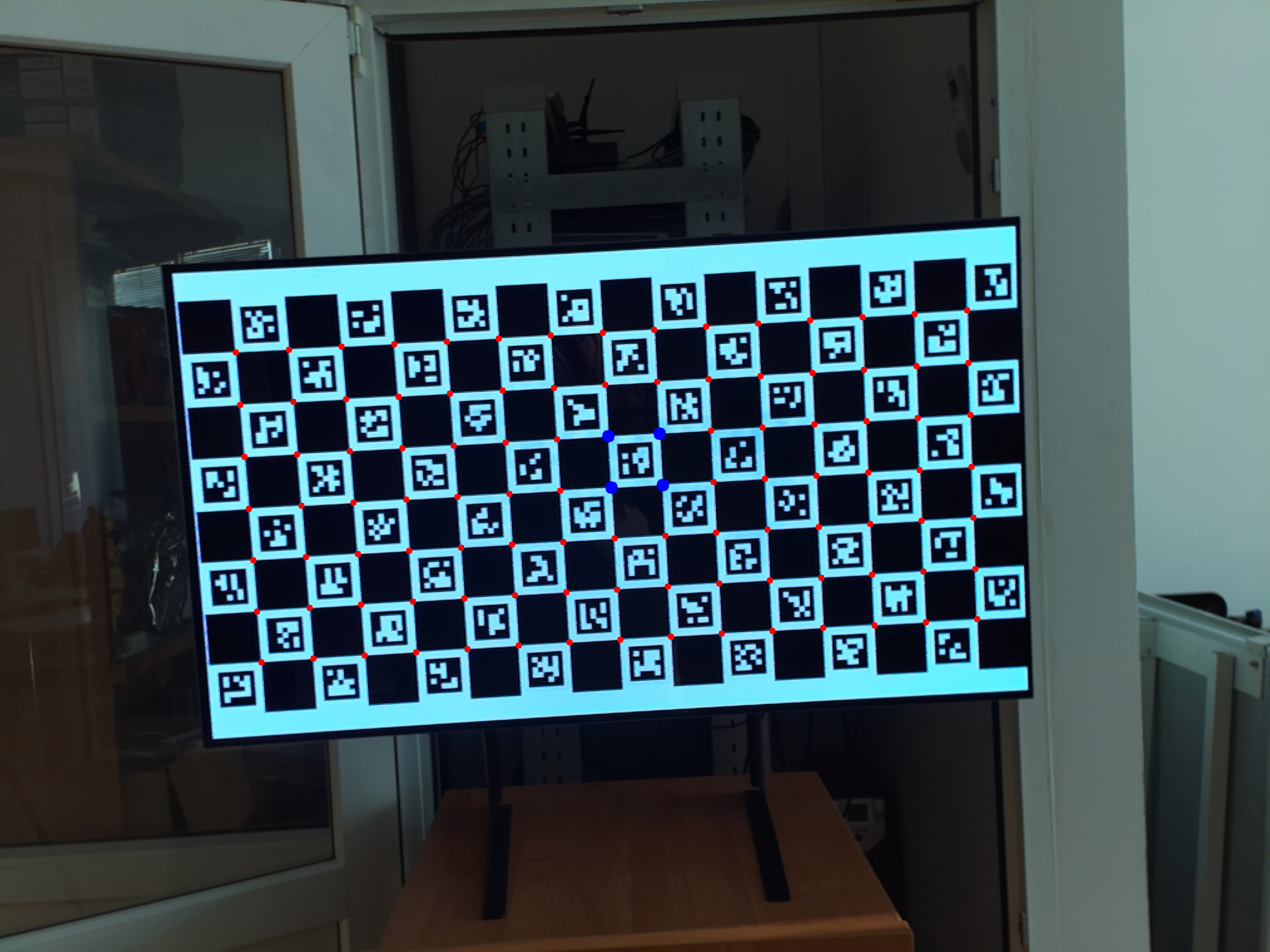}
&
\includegraphics[width=0.48\linewidth]{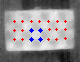}
\\
\includegraphics[width=0.48\linewidth]{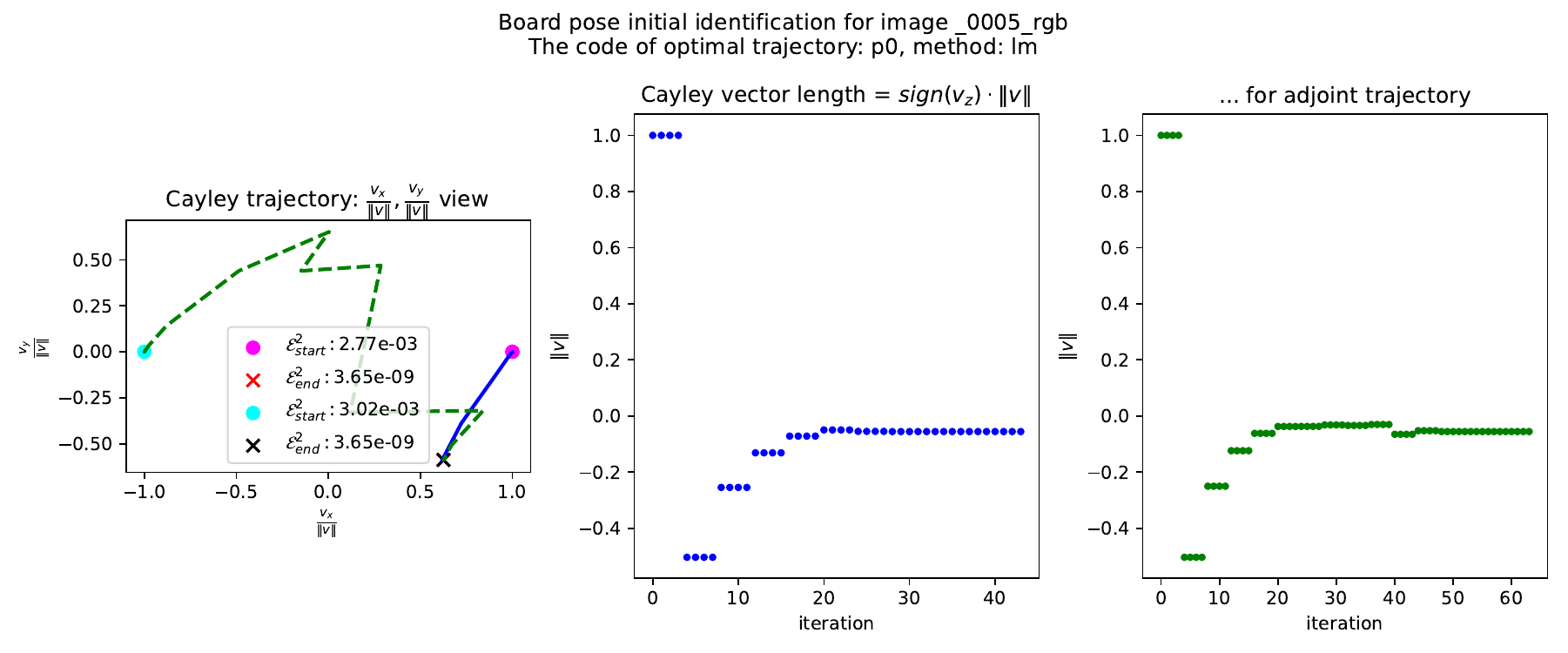}
&
\includegraphics[width=0.48\linewidth]{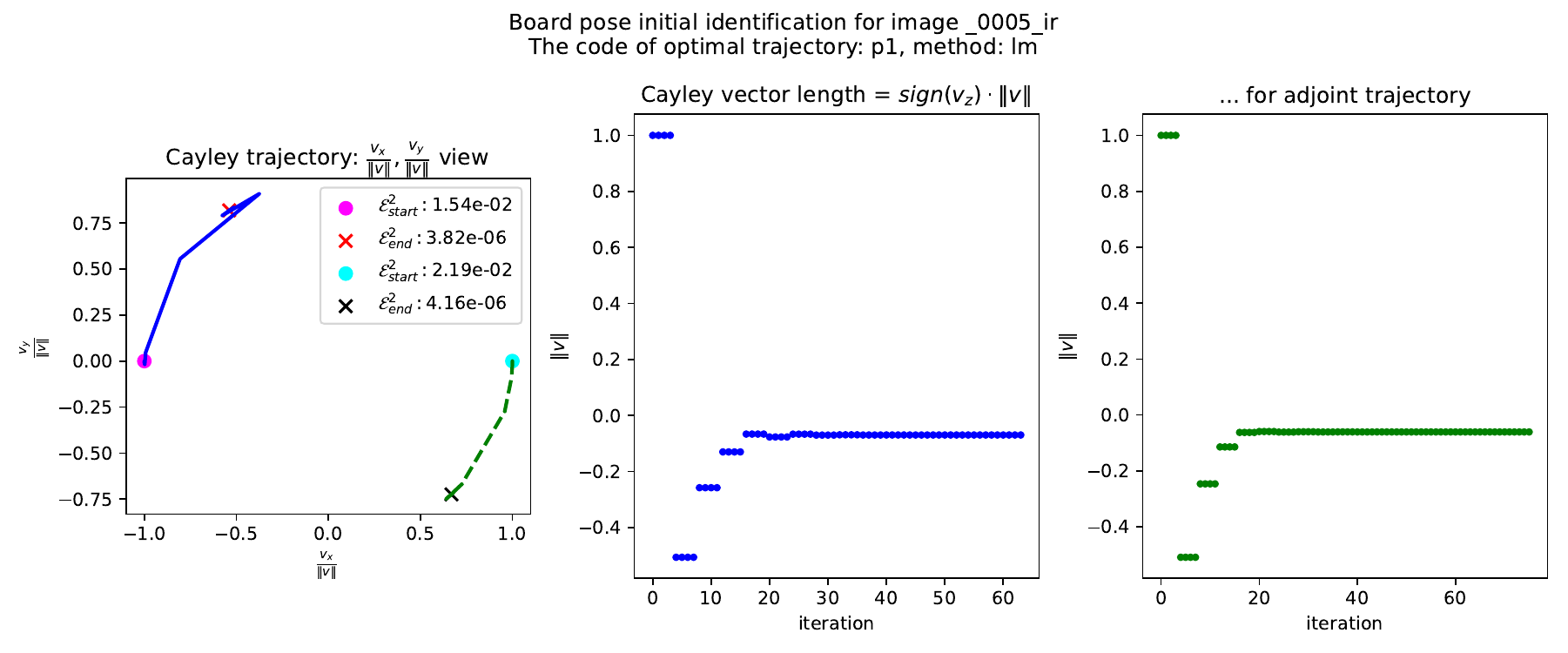}
\end{tabular}
\caption{
Calibration board views (id=05) displayed sequentially on an OLED screen: on the left, a board with Charuco markers; on the right, a thermal image of a standard chessboard.
The chessboard grid perfectly overlaps the Charuco grid at half the resolution. In the second row, trajectories of the LM optimizer states in Cayley space are shown for RGB camera pose estimation (left) and IR camera pose estimation (right), respectively.
}
\label{fig:traj-05a}
\end{figure}

\begin{figure}[H]
\begin{tabular}{c}
\includegraphics[width=0.99\linewidth]{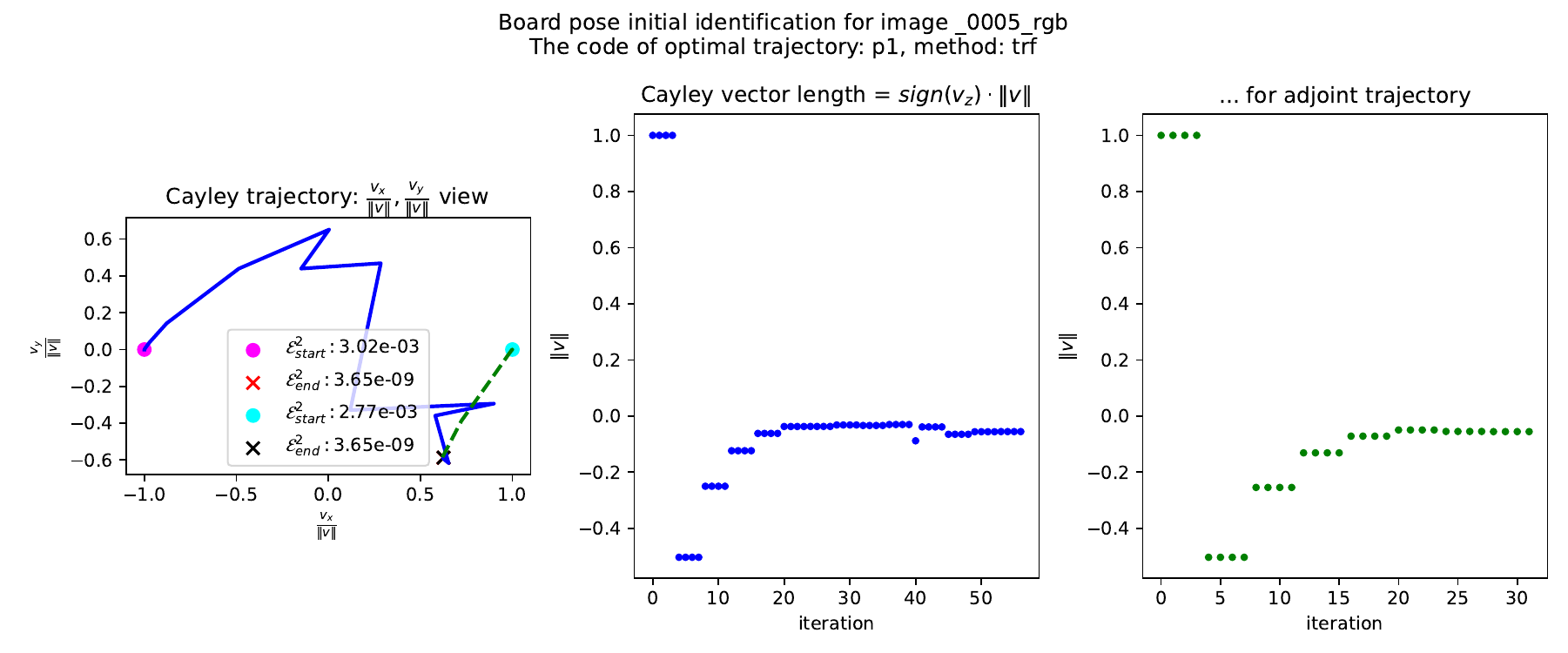}
\\
\includegraphics[width=0.99\linewidth]{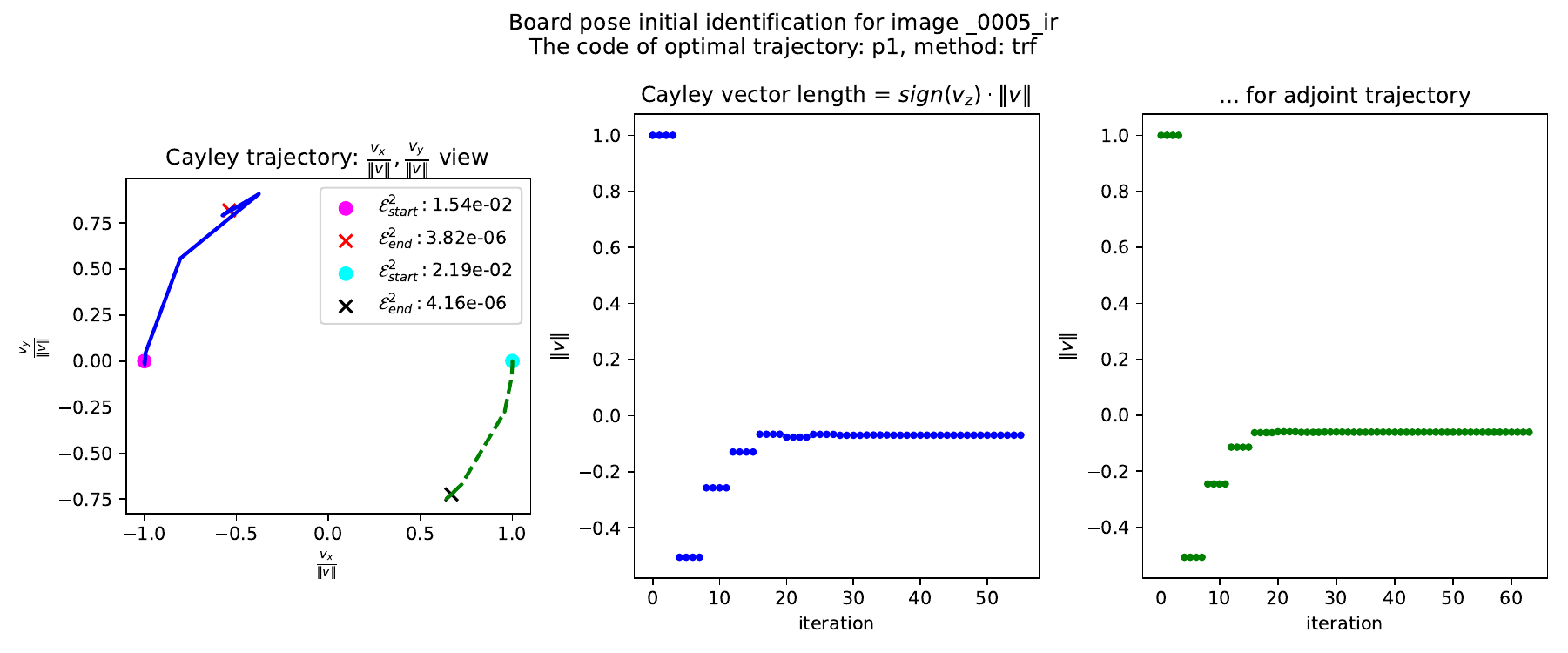}
\end{tabular}
\caption{
Trajectories of the TRF optimizer states in Cayley space.
The first row corresponds to RGB camera pose estimation,
while the second row shows the trajectory for the low-resolution thermal camera.
}
\label{fig:traj-05b}
\end{figure}

\begin{figure}[H]
\begin{tabular}{cc}
\includegraphics[width=0.48\linewidth]{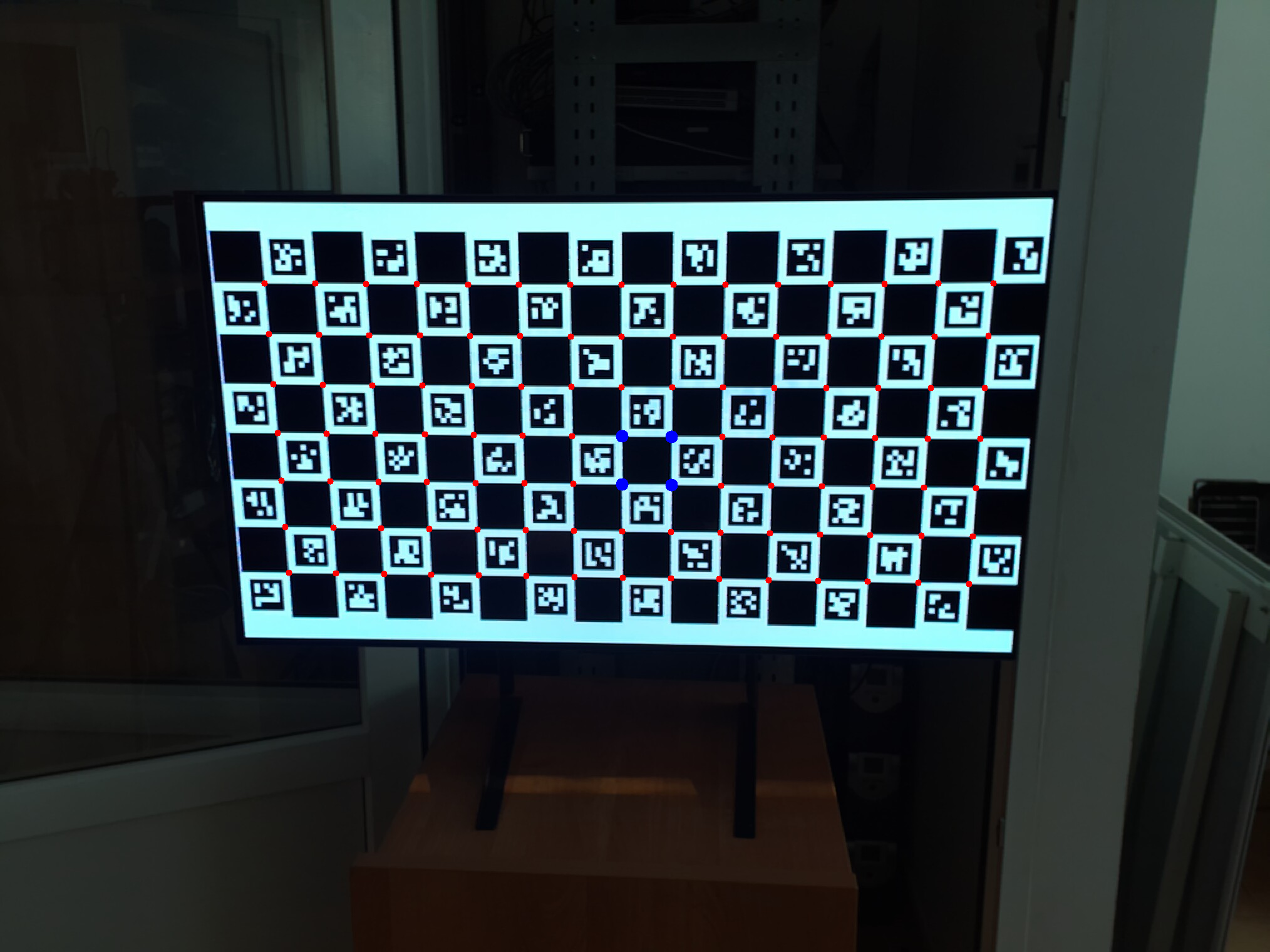}
&
\includegraphics[width=0.48\linewidth]{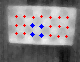}
\\
\includegraphics[width=0.48\linewidth]{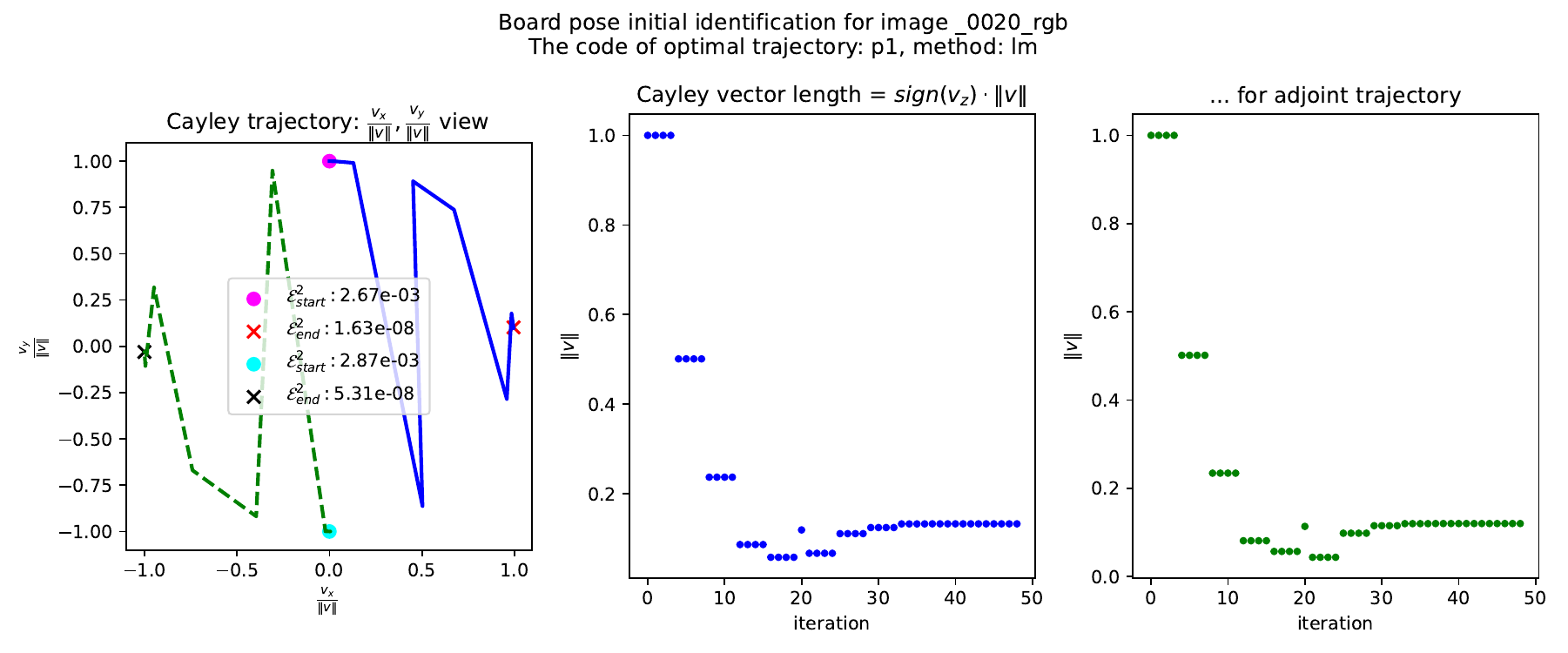}
&
\includegraphics[width=0.48\linewidth]{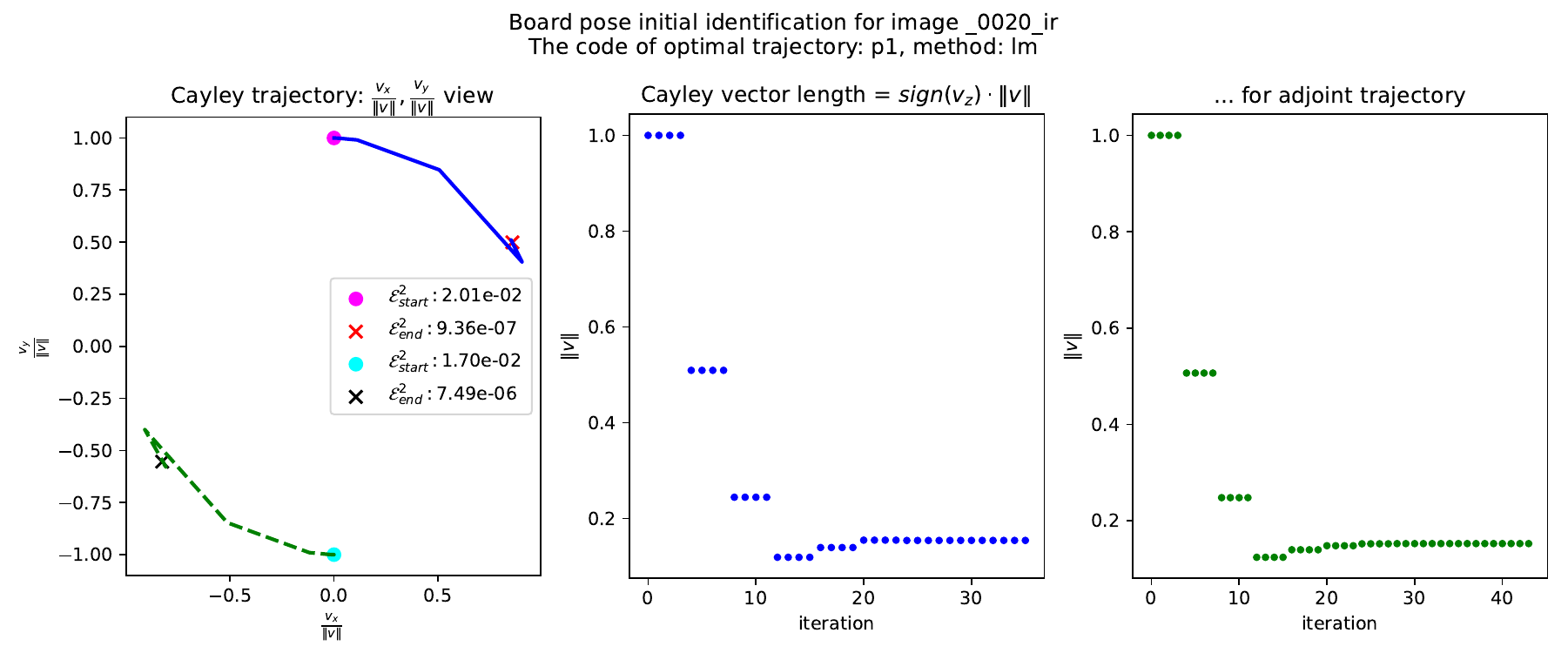}
\end{tabular}
\caption{
Calibration board views (id=20) displayed sequentially on an OLED screen: on the left, a board with Charuco markers; on the right, a thermal image of a standard chessboard.
The chessboard grid perfectly overlaps the Charuco grid at half the resolution. In the second row, trajectories of the LM optimizer states in Cayley space are shown for RGB camera pose estimation (left) and IR camera pose estimation (right), respectively.
}
\label{fig:traj-20a}
\end{figure}

\begin{figure}[H]
\begin{tabular}{c}
\includegraphics[width=0.99\linewidth]{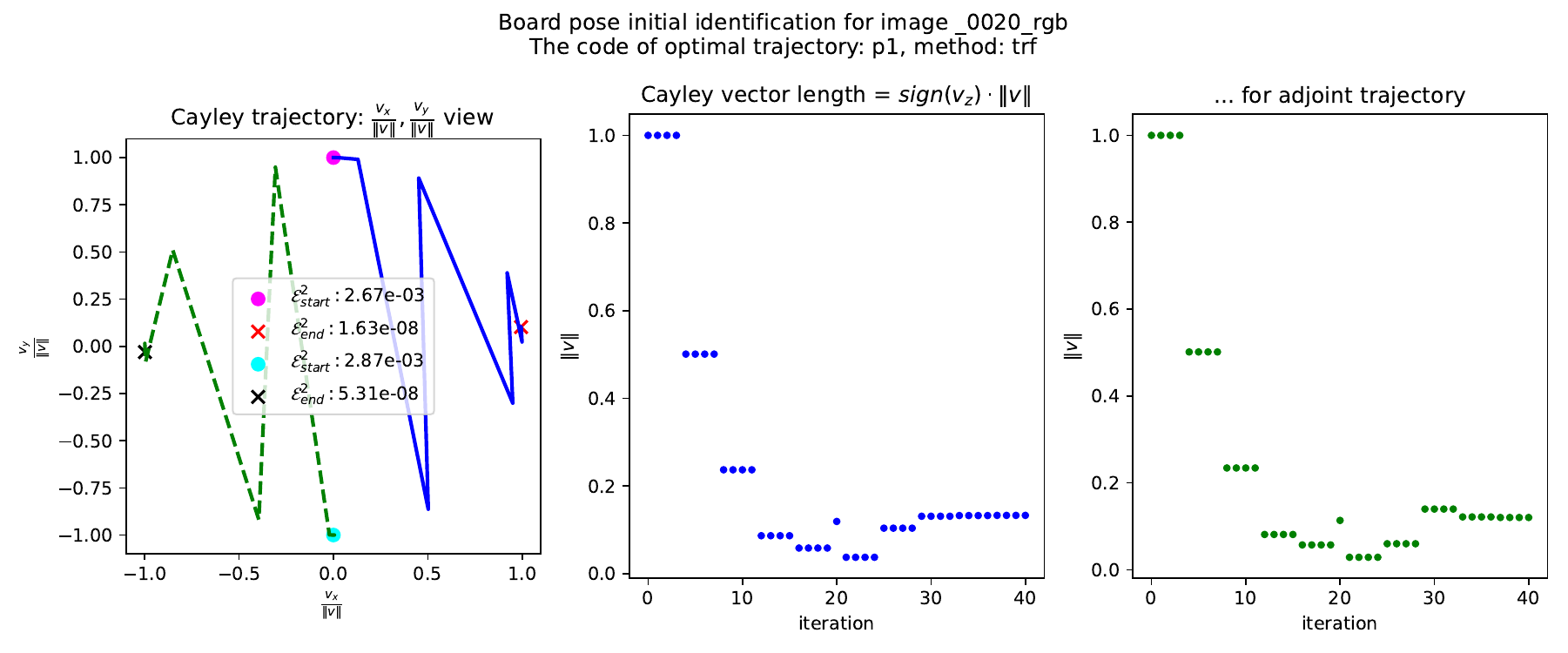}
\\
\includegraphics[width=0.99\linewidth]{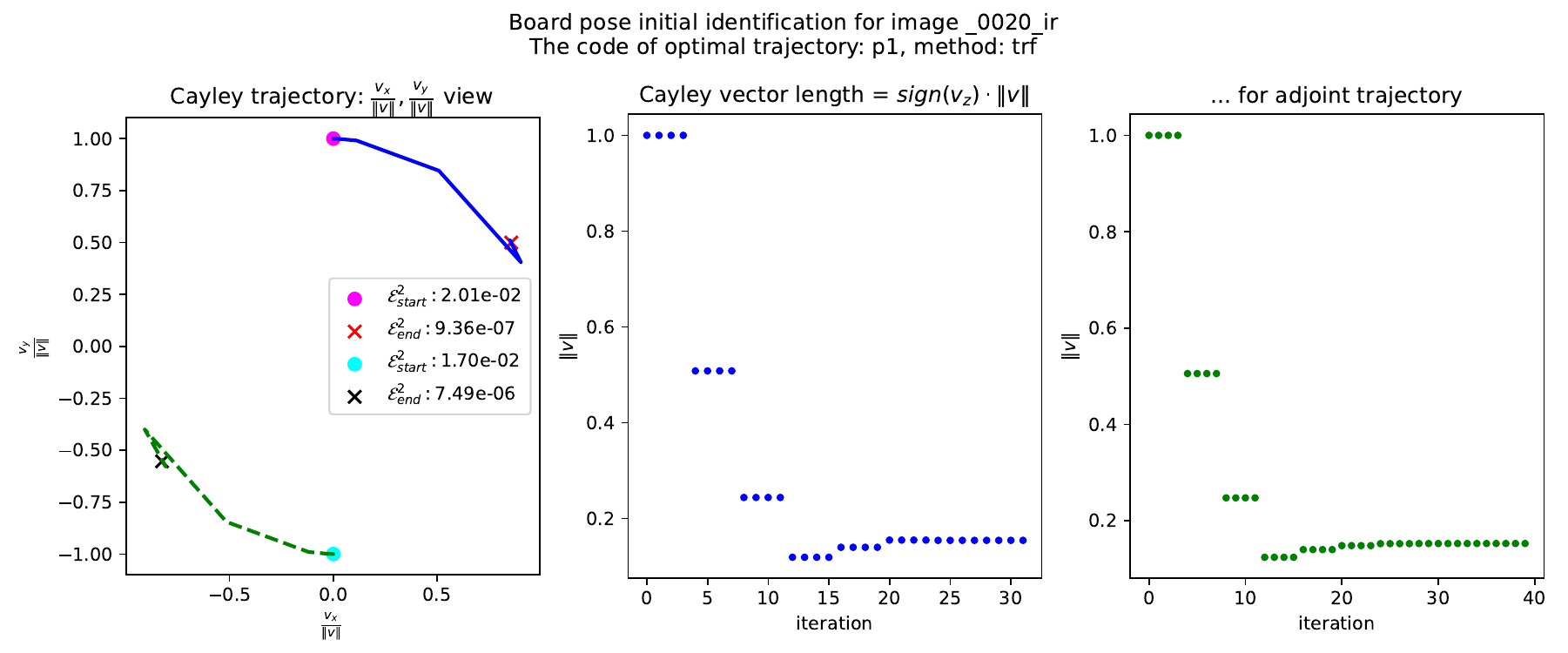}
\end{tabular}
\caption{
Trajectories of the TRF optimizer states in Cayley space.
The first row corresponds to RGB camera pose estimation,
while the second row shows the trajectory for the low-resolution thermal camera.
}
\label{fig:traj-20b}
\end{figure}

\end{document}